\documentclass[journal]{IEEEtran}
\usepackage{amsmath,amsfonts}
\usepackage{algorithmic}
\usepackage{algorithm}
\usepackage{array}
\usepackage[caption=false,font=normalsize,labelfont=sf,textfont=sf]{subfig}
\usepackage{textcomp}
\usepackage{stfloats}
\usepackage{url}
\usepackage{verbatim}
\usepackage{enumitem}
\usepackage{graphicx}
\usepackage{cite}
\usepackage{multirow}
\usepackage{tabularx}

\usepackage{xcolor}

\begin{document} 

\title{Modeling, Embedded Control and Design of Soft Robots using a Learned Condensed FEM Model}

\author{Tanguy Navez$^{*1}$, Etienne Ménager$^{*2}$, Paul Chaillou$^{1}$, Olivier Goury$^{1}$, \\ Alexandre Kruszewski$^{1}$ and Christian Duriez$^{1}$  

\thanks{$^{*}$ Tanguy Navez and Etienne Ménager contributed equally to this work. $^{1}$Univ. Lille, Inria, CNRS, Centrale Lille, UMR 9189 CRIStAL, F-59000 Lille, France. $^{2}$Inria, Département d’informatique de l’ENS, Ecole normale supérieur, CNRS, PSL Research University, Paris, France. 
        {\tt\small etienne.menager@inria.fr, \tt\small tanguy.navez@inria.fr}}%
}


\maketitle

\begin{abstract}

The Finite Element Method (FEM) is a powerful modeling tool for predicting soft robots' behavior, but its computation time can limit practical applications. In this paper, a learning-based approach based on condensation of the FEM model is detailed. The proposed method handles several kinds of actuators and contacts with the environment. We demonstrate that this compact model can be learned as a unified model across several designs and remains very efficient in terms of modeling since we can deduce the direct and inverse kinematics of the robot. Building upon the intuition introduced in~\cite{Menager2023}, the learned model is presented as a general framework for modeling, controlling, and designing soft manipulators. First, the method's adaptability and versatility are illustrated through optimization-based control problems involving positioning and manipulation tasks with mechanical contact-based coupling. Secondly, the low-memory consumption and the high prediction speed of the learned condensed model are leveraged for real-time embedding control without relying on costly online FEM simulation. Finally, the ability of the learned condensed FEM model to capture soft robot design variations and its differentiability are leveraged in calibration and design optimization applications.

\end{abstract}


\section{Introduction}

Compared to rigid robots, soft robots are made with low-stiffness materials, enabling them to adapt to and interact with their environment~\cite{Laschi2016}. These features make them ideal for various applications, such as navigating confined spaces, handling fragile objects, interacting with biological tissues, or solving complex tasks using contacts. However, this paradigm introduces greater complexity due to the strong interplay between soft materials, geometry, and actuation. As a result, modeling, control, and design become more challenging, requiring the development of specialized tools. 
In this work, a general framework for the control and design of soft manipulators based on the learned condensed FEM model is built upon the idea proposed in~\cite{Menager2023}. This section contextualizes the condensed FEM model in the soft robot simulation, model reduction, and design bibliographies.

\subsection{Simulation, modeling and control of soft robots }

Physics-based simulation of soft robots is a safe, fast, and cost-effective way to test designs and control strategies~\cite{Choi2020}. The simulation must be sufficiently realistic to accurately capture the robot's behavior and its interactions with the environment. Numerous models exist for simulating soft robots~\cite{armanini_soft_2022}.

Geometric models, such as the Piece-wise Constant Curvature (PCC) model~\cite{Webster2010}, can be used to describe soft robots under strong geometrical assumptions. PCC assumes a constant curvature for soft robot shapes composed of beam elements. However, this assumption becomes invalid when external forces, such as gravity or contacts, are applied, making PCC unsuitable for these cases. Other heuristic models include pseudo-rigid models~\cite{Khoshnam2013}, which represent the robot as a series of rigid links connected by joints, and mass-spring models~\cite{Lloyd2007}, where the robot is modeled as masses connected by spring networks. These models are not directly linked to the robot’s mechanical properties,  requiring specific optimization methods for effective real-world application.
The use of more complex models, such as those based on continuum mechanics equations or neural networks, may be necessary to capture complex non-linear mechanical behavior. Numerical methods based on continuum mechanics are slower than analytical solutions but allow for the simulation of soft robots with fewer assumptions about geometry or behavior. However, the more accurate the simulation, the more computationally intensive it becomes due to non-linearities and fine discretization. For many soft robotics applications involving slender structures, models like the Euler-Bernoulli~\cite{EulerBernouilli2015} rod or the Cosserat~\cite{Cosserat2018} beam can be solved using numerical methods. A common limitation of these approaches is that they assume (almost) rigid cross-sections, meaning all deformations are not taken into account. In some cases, this may not accurately capture the behavior of certain materials, large transverse deformations resulting from actuation strategies like pneumatic cavities, or specific robot configurations.
Generic methods can help overcome these limitations. The Material Point Method (MPM) \cite{Zhang2015} is a mesh-free approach representing a continuum body with multiple small Lagrangian elements. However, MPM is more memory-intensive and slower than other numerical methods based on continuum mechanics equations. As a result, applications of MPM to soft robots in the literature typically focus on controlling or optimizing simplified robot models~\cite{Hu2018ChainQueenAR}.
FEM methods rely on spatial discretization of the robot's geometry using a mesh. With appropriate mesh resolution, FEM methods can simulate a wide variety of robots with high precision. While FEM methods are more efficient for complex robot simulations in terms of memory and computation time~\cite{RodriguezRealTime2017}, their application in real-time scenarios remains constrained by computational resources, especially when dealing with dense meshes or complex boundary conditions, such as contact interactions.
Finally, neural network models enable robot behavior to be learned from experimental data\cite{ThuruthelLearnedModel2017, GillespieLearningNonLinear2018, BernLearnModel2020}. However, they often require large amounts of training data to be accurate, which can limit their use in situations where data is scarce or expensive to collect.

The models chosen to describe the behavior of soft robots influence their control strategies. For instance, due to their small dimension, beam models can be used for fast control loops~\cite{LiPLSC2023} or predictive control~\cite{SpinelliMPC2022}.  However, their one-dimensional nature makes them unsuitable for handling complex robot geometries. Other simplified models, such as the voxel-based mass-spring model used in EvolutionGym~\cite{BhatiaEvolGym2021}, are computationally efficient and allow for easy exploration of complex tasks, but they do not transfer well to physical robots. Learned models have been applied to low-level adaptive controllers~\cite{ThuruthelLearnedModel2017, ThuruthelStableOpenLoop2018}, but their performance is limited to the training data set. Finally, models based on 3D continuum mechanics can provide fine control of robot deformations. For example, the FEM model can be condensed by projection into a smaller space~\cite{duriez2013control}, but this still requires computing the full FEM model. The linearization of the control command and the dimension of the related mathematical systems shorten the control horizon and limit control to low-level tasks.
The introduced models can also be used to support Reinforcement Learning (RL) algorithms for optimizing controllers for high-level tasks. For instance, Chainqueen~\cite{Hu2018ChainQueenAR} and DiffTaichi~\cite{Difftaichi} have been applied to simulate soft robots and learn control strategies. However, these simulators use simplified models and do not offer a wide range of actuation methods, deformable structures, or material laws. SofaGym~\cite{SofaGym}, a RL framework based on FEM simulation, enables solving complex tasks with accurate robot models. Nevertheless, learning control strategies requires extensive interaction between the robot and its environment, which becomes time-consuming when the simulation is slow.

\subsection{Model Order Reduction for Soft Robotics}

Using reduced models is a common approach in soft robotics to simplify the modeling and control of these complex systems. Model order reduction techniques aim to decrease the computational complexity of high-dimension mathematical models in numerical simulations. These reduced models are created by reducing the system's degrees of freedom and can be derived through different methods, such as projection, simplification, or learning.

Mixed methods like Proper Orthogonal Decomposition (POD) can be applied to reconstruct a robot's state based on its deformation modes~\cite{Goury2018}, typically focusing only on the most significant modes. This approach requires an offline learning phase and is coupled with condensed modeling and additional control tools from~\cite{Coevoet2019} for real-time inverse control of soft robots. However, using this reduced model for embedded control applications remains computationally intensive. Koopman filters~\cite{Koopman2022} can also capture important system features, such as deformation modes, but the accuracy of the linear representation depends on the choice of the observation functions.
Simplifying the robot's mechanical behavior can be done using beam models like Euler-Bernoulli or Cosserat, which represent the behavior of more complex robots within a lower-dimensional space~\cite{proxy2023}. However, these simplified models only capture the robot's behavior under specific assumptions, such as the characteristics of contact interactions.
The model reduction method by condensation involves projecting the system onto a lower-dimensional subspace~\cite{Menager2023}. This technique is used in FEM to control soft robots through inverse modeling and optimization~\cite{Coevoet2019}.
Learning methods enable the estimation of reduced models from experimental data. Various approaches can be used. For example, a data-driven approach was applied to learn a differentiable inverse model of a soft robot’s quasi-static behavior, linking actuation to end-effector positions~\cite{Bern2020}. However, since this direct link is task-specific, the learned reduced model is only applicable under the conditions in which the data was collected.
Each scenario requires specific training. Additionally, learned models may not capture the transient dynamic effects of the system, potentially impacting the performance of the associated controller. Other learning-based methods enable multi-time-step control. For example, a spectral sub-manifold can be used to learn the robot's dynamics around an equilibrium point~\cite{Alora2023}. However, this method relies on approximate dynamics and is limited to robots with equilibrium points and no contact points.

Using reduced models significantly affects system controllability. Simplifying the model decreases controller complexity, making its design more straightforward and efficient. In addition, reduced models facilitate faster and more efficient simulation, which are valuable for designing and optimizing reactive controllers~\cite{ThuruthelLearnedModel2017, LiPLSC2023}. However, this simplification can also present challenges. Reducing the number of degrees of freedom may result in the loss of critical information about the system behavior, impacting model accuracy and controller performance. For instance, in the proxies method~\cite{proxy2023}, a simplified model has to be combined with a FEM-based control loop to correct errors from the beam-based model. Moreover, reduced models might not capture non-linear effects and complex environmental interactions, limiting their use in more complex scenarios. This is particularly true for models learned from a specific set of contacts~\cite{Goury2018}, which are often difficult to generalize to different contact conditions.

\subsection{Design in Soft Robotics}

For a long time, soft robot designs were inspired by morphological concepts observed in nature~\cite{Bioinspiration2013}. However, directly mimicking biological organisms has limitations. Fabricating robots at the complexity level of biological organisms is currently unachievable, and the environments and tasks for robotics applications often do not replicate those in nature. Given the vast design spaces encountered in soft robotics, automatic computational design methods have emerged to facilitate the design process~\cite{PinskierAuto2022}. Designing soft robots is complex due to the intricate coupling between the material properties, geometry, and soft body dynamics. Both material and geometry must be optimized based on the task requirement, such as force, deformation, and speed.

Model-based design methods rely on specific strategies to leverage information from soft robot models. These methods require writing objectives function and analytical gradient with respect to continuous design variables. 
For example, frameworks have been developed to optimize the fiber orientation of fiber-reinforced pneumatic actuator~\cite{connolly_automatic_2017} or the placement of cable actuators~\cite{skouras_computational_2013} for achieving different deformation states.
Topology optimization, a numerical technique that redistributes material within a specific design domain, has been widely used to generate soft robot structures~\cite{PinskierAuto2022} and has recently been extended to generate multi-materials soft robot~\cite{PinskierVoxelTopo2024}. While these methods benefit from fast convergence due to first-order optimization techniques, they are limited to static conditions and small displacements. The effectiveness of model-based design methods depends on the user’s ability to create mathematical formulations that accurately reduce the feature space while maintaining high accuracy. These approaches are tailored to specific scenarios with particular constraints, making it challenging to adapt them to different design variables and scenarios, especially for non-specialists in mechanical modeling and mathematical programming. Additionally, since these methods do not account for interactions with the environment, they are limited to soft components or soft robots without direct interaction with their surroundings.

Another approach is to use model-free methods that treat simulation as a black box to evaluate design performances, employing algorithms such as sensitivity analysis or stochastic methods to explore the design space. These methods are easier to implement since they do not require rewriting specific soft robot models relative to design variables. For example, Bayesian or evolutionary algorithms have been used jointly with FEM simulation to optimize the design of a pneumatic actuator~\cite{dammer_polyjet-printed_2019} and a soft gripper~\cite{NavezDesignOptiContact2024} with self-contacts. These approaches show great promise for transferring numerically optimized results to physical prototypes, leveraging accurate FEM simulation~\cite{NavezDesignOptiContact2024, automaticdesign}. However, they often suffer from low sample efficiency, leading to high computational costs related to design performance assessment in simulation.

A trend is emerging towards using surrogate models to facilitate the exploration of soft robot design. These models provide approximate representations of system behavior under different working conditions and have been applied to optimize robot design according to specific performance criteria. For instance, in~\cite{Yao2023}, a surrogate model is learned to predict the angular deflection of pneumatic modules as a function of mechanical and geometric parameters. However, changing the cost function in this approach requires generating new data and relearning the surrogate model. In~\cite{Auxetic2023}, a FEM-based differentiable surrogate model is used for joint optimization of the design and trajectory of an auxetic robot. This model predicts the behavior of each lattice node and is limited to cylindrical geometries, restricting its applicability to other design parameters. Finally, a soft crawler robot's leg placement has been optimized to maximize locomotion capacity~\cite{automaticdesign}, relying on pre-learned POD-based reduced models of the legs as presented in~\cite{Goury2018} to speed up the simulation. This application is particularly compelling as it co-optimizes both the morphology and control of the soft robot, enabling the emergence of Embodied Intelligence.
However, not all reduced models are suitable for design optimization. For example, the POD-based model in~\cite{Goury2018} is mesh- and design-dependent, making it difficult to link explicitly to geometric design parameters. In the case of the soft crawler robot optimization~\cite{automaticdesign}, this limitation prevents consideration of variations in leg geometry. Moreover, other learning-based models presented in \cite{ThuruthelLearnedModel2017, Alora2023} do not explicitly depend on the design of the soft robots.

\begin{figure*}[!ht]
\centering
\resizebox{0.85\textwidth}{!}{\includegraphics{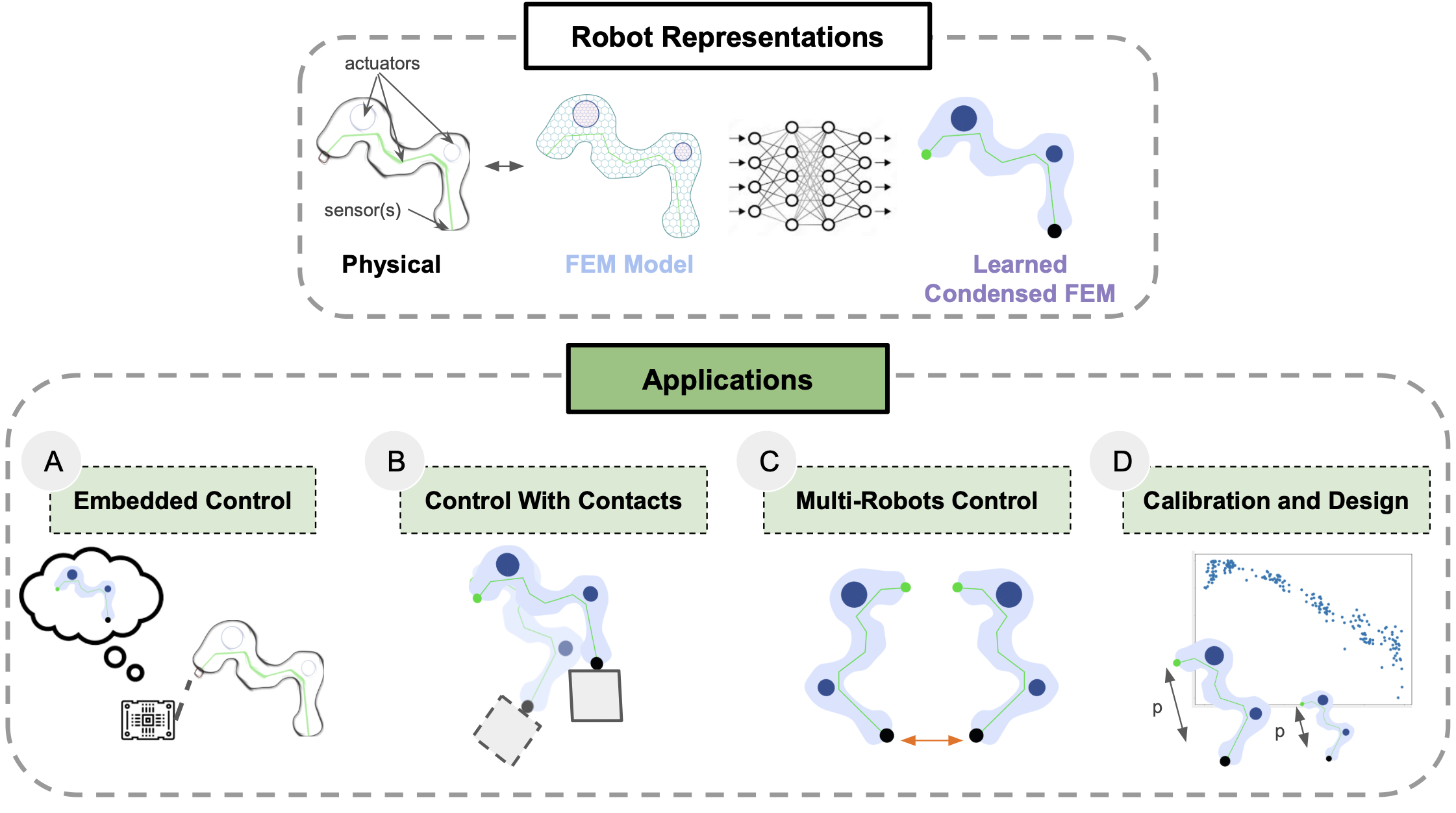}}
\caption{Illustration of the proposed framework and its applications. The FEM model of a robot is projected in the constraint space, and the corresponding matrices are learned using a neural network. The learned matrices can be used in different applications like A) real-time embedded control, B) inverse control involving predefined contact points, C) control of multiple identical robots from a single learned model, and D) both design optimization and calibration applications.}
\label{fig:Intro_CondensedFemFramework}
\end{figure*}

\subsection{Contributions}

In this paper, we propose a new method for building a reduced model of soft robots without making assumptions about their shape. This model captures both the inverse and direct kinematics of the robot under quasi-static assumptions and accounts for contact configurations. Compared to other reduced models, our method uses machine learning to obtain a compact mechanical representation of a soft robot based on a FEM model that is not specific to a single task.

We combine both learned and computed models, leveraging the speed and flexibility of learned models and the precision and formalism of computed ones. The goal is to quickly predict mechanical quantities that can be used to define different design or control optimization problems. Unlike models that learn input-output relationships or use auto-encoders, we focus on learning mechanical quantities derived from continuum mechanics. These quantities enable the use of low-level controllers based on inverse robot modeling and optimization tools. In addition, we demonstrate that our reduced model is compact enough to run on a microcontroller. 

Finally, the proposed learned condensed FEM model can be used as a surrogate model to solve design tasks. It offers a general approach to capture both the design and control of soft robots.  Unlike current methods, the model doesn't need to be relearned when the objective changes. Indeed, the mechanical matrices predicted by the model can be used to define various design objectives. As a result, multiple robot design variations can be controlled using a single learned condensed FEM model, providing the opportunity for Embodied Intelligence applications.

The proposed general framework builds upon the approach introduced in~\cite{Menager2023}. Our contributions, illustrated in Figure~\ref{fig:Intro_CondensedFemFramework}, are as follows:
\begin{itemize}
    \item We evaluate the low memory usage of the condensed FEM model and experimentally validate its effectiveness for embedded control of a pneumatic soft robot (Figure~\ref{fig:Intro_CondensedFemFramework}.A).
    \item We extend the model to handle contact management (Figure~\ref{fig:Intro_CondensedFemFramework}.B). We highlight the modularity of the condensed FEM model by using it to control a complete Soft Gripper in a manipulation task, based on the model learned from a single Soft Finger (Figure~\ref{fig:Intro_CondensedFemFramework}.C). 
    \item We extend the model to manage design parameters. We demonstrate its independence from the mesh size and its inherent differentiability in both calibration and design optimization applications (Figure~\ref{fig:Intro_CondensedFemFramework}.D).
\end{itemize}

We illustrate these contributions using different robots. Both our study and the previous work~\cite{Menager2023} show that the method can be applied to various robot types (parallel or continuous) with different types of actuation (cable or pneumatic). The robots chosen for this article allow for comparison with a baseline. For the Stiff-Flop robot (see section \ref{sec:embeddedControl}), we already have a reduced model, but it is still too slow for embedded control. For the Finger robot (see sections \ref{sec:controlWithContact} and \ref{seq:designOptim}), parametric approaches have already been proposed to address design issues. Our goal is to present a method that can handle both control and geometric/mechanical variations with a single model. The Finger robot can also serve as a functional manipulator for object manipulation. However, the method is general and can be applied to other types of robots, as long as they meet the model's assumptions.
 
\section{Soft Robot FEM Modeling}
\label{sec:softrobotmodeling}

This section introduces the finite element framework used for modeling soft robots.
This framework is general, allowing for any soft robot shape, material laws, and constraints. However, robots are represented by a set of high-dimensional matrices, with their dimension directly proportional to the number of mesh elements. To reduce the FEM model dimension, these matrices can be projected into the constraint space for direct and inverse modeling of soft robots.

\subsection{Quasi-Static FEM Modeling}

The mechanical behavior of soft robot deformations is described using continuum mechanics, which lacks analytical solutions in the general case. Non-linear FEM is one of the numerical methods employed to obtain a converging approximate solution. In this paper, the quasi-static formulation is considered. Based on solid mechanics, the quasi-static equilibrium is written:

\begin{equation}
\begin{aligned}
     K(x) \Delta x = f_{\text{ext}} - f_{\text{int}}(x)
     + H_{a}^T \lambda_{a} + H_{c}^T \lambda_{c}
\end{aligned}
\label{eqQuasiStaticEquilibrium}
\end{equation}

where $\Delta x$ are small displacement of nodes around their current position $x$, $K(x)$ the tangent stiffness matrix depending on the current positions of the FEM nodes. Actuation and contacts are modeled as constraints, using respectively Lagrange multipliers $\lambda_{a}$ and $\lambda_{c}$. This is convenient for both forward and inverse modeling. The corresponding Jacobians, in the mathematical sense, are respectively $H_{a}$ and $H_{c}$. The external forces, such as gravity, are represented by $f_{ext}$, while $f_{int}$ represents the non-linear internal forces of the deformable structure computed from the material laws. In this work, we consider the FEM model with geometric non-linearity (large displacements) but linear elasticity (small strain). The model is characterized by a Poisson ratio $\nu$ and a Young Modulus $E$. However, the framework should be extended to hyper-elastic models with large strain.  

\subsection{Condensed FEM Modeling}

Soft robot motion is controlled through specific points called effectors. The most common approach is to control their position by minimizing the squared norm of the shift $\delta_{e}(x) = x_{\text{effector}} - x_{\text{goal}}$ between the effectors and the goals. Three kinds of constraints are considered: actuation constraints $a$, contact constraints $c$, and effector constraints $e$. The first two are considered as active constraints, while the last one is passive. Each constraint is associated with a Jacobian matrix $H_j$ and a characteristic distance $\delta_j$ for $j \in \{a, e, c\}$. $\delta_a$ represents a displacement, which depends on the specific actuator model. In the literature, $\delta_c$ usually refers to the signed distance between the contact points of two colliding objects. In this paper, since there is no prior knowledge about the object colliding with the robot, $\delta_c$ is defined as the distance between the contact points and their rest positions.

From one computation step to another, the constraint equations are linearized thanks to their Jacobian
\begin{equation}
     \forall{j \in \{a, e, c\}}:  \ \delta_j(x_{i}) = \delta_j(x_{i-1}) + H_j \Delta x_i 
\label{eqQuasiStaticFreeEquilibrium}
\end{equation}

The constraint is resolved in two steps by decomposing $\Delta x$ from Equation~\ref{eqQuasiStaticEquilibrium} as follows:

\begin{equation}
\begin{aligned}
\left\{
\begin{array}{l}
     K(x_{i-1}) \Delta x^{\text{free}}_{i} = f_{\text{ext}} - f_{\text{int}}(x_{i-1}) \\
     K(x_{i-1}) \Delta x^{\text{const}}_{i} = H_{a}^T \lambda_{a} + H_{c}^T \lambda_{c}
\end{array}
\right.
\end{aligned}
\label{eqQuasiStaticFreeEquilibrium}
\end{equation}

Solving the first equation for $\Delta x^{\text{free}}$, allows evaluating $\delta_{j}^{\text{free}}$ $j \in \{a, e, c\}$ using $\delta_{j}^{\text{free}} = \delta_j(x_{i-1}) + H_i \Delta x^{\text{free}}_{i}$.

With the projection of the compliance in the constraint space expressed as $W_{jk} = H_j K^{-1}(x_{i-1}) H_k^T \text{ for } j,k \in {a, e, c}$, this results in:

\begin{equation}
\begin{aligned}
    \delta_{a}(x_{i}) = W_{aa} \lambda_a +  W_{ac} \lambda_c + \delta_{a}^{\text{free}} \\
    \delta_{e}(x_{i}) = W_{ea} \lambda_a +  W_{ec} \lambda_c + \delta_{e}^{\text{free}} \\
    \delta_{c}(x_{i}) = W_{ca} \lambda_a +  W_{cc} \lambda_c + \delta_{c}^{\text{free}} \\
\end{aligned}
\label{eq:schur}
\end{equation}

The matrices $W_{ij}$ are the Schur complements~\cite{Covoet2019These}, representing the mechanical coupling between actuators $a$, effectors $e$, and contacts $c$. The $\delta_j \text{ for } j \in \{a,e,c\}$ are computed from the same distance vector $\delta_j^{\text{free}}$ evaluated at the previous equilibrium.

Both $W_{ij} = H_i K^{-1}(x) H_j^T \text{ for } i,j \in {a, e, c}$ and $\delta_{j}^{\text{free}}$ $j \in \{a, e, c\}$ can be seen as a reduced model of the robot, as illustrated in Figure~\ref{fig:robots_contact}.

\subsection{Direct and Inverse Modeling}

In the direct problem, when the intensity of the actuator forces $\Delta \lambda_{a}$ changes the corresponding motion of the effectors $\Delta \delta_{e} = W_{ea} \Delta \lambda_{a}$ can be computed. Since actuators are also controlled in displacement, changing the actuators' position by $\Delta \delta_{a}$ leads to the effectors' motion through the equation $\Delta \delta_{e} = W_{ea} W_{aa}^{-1} \Delta \delta_{a}$. This equation provides the Jacobian $ J = W_{ea} W_{aa}^{-1}$, which represents the direct kinematics of the soft robot~\cite{zhang2016kinematic}. 

In the inverse problem, the actuator constraints are not known in advance, requiring quadratic optimization problems to be solved for the actuator forces using dedicated solvers \cite{Coevoet2019}. Solving the inverse problem is about finding the solution to the following optimization problem: 

\begin{equation}
    \begin{aligned}
     arg_{\lambda_a}&\text{min} \quad ||\delta_{e}||^2 \\
    s.t. &  \delta_{\text{min}} \leq \delta_a \leq \delta_{\text{max}} \text{ (Actuators course constraint)}  \\ 
    &  \lambda_{\text{min}} \leq \lambda_a \leq \lambda_{\text{max}} \text{ (Actuation effort constraint)}  \\ 
    & 0 \leq \delta_c \perp \lambda_c \geq 0 \text{ (Contact handling)}
    \end{aligned}
\label{eq:QPActuationContacts}
\end{equation}

where the responses to the contact constraints are calculated using both Signorini's law and Coulomb's law. A comprehensive overview of the collision pipeline is provided in~\cite{SOFA}. 

This background shows that both direct and inverse models of soft robots can be derived, as long as approximations of Equations \ref{eq:schur} giving the effective state of the actuation $\delta_{a}$ and the contact $\delta_{c}$ are available.

\section{Learned Condensed FEM Modeling}

This section introduces the learned condensed FEM model. We explain why this model was chosen, how it is trained, and how it is applied in control applications.

\subsection{Mechanical Representation Learning}

In this work, a high-dimensional state of the robot is available through simulation. To reduce the dimension of this state, the global compliance matrix projected onto the constraint space $W$ (see section \ref{sec:softrobotmodeling}) is a good candidate. This is because $W$ is a low-dimensional matrix, independent of the FEM mesh size, and it directly links the actuator and effector constraint spaces. Combined with the free displacement vector $\delta^{\text{free}}$, $W$ can be effectively used to model and control a soft robot.

\begin{figure*}[!ht]
\centering
\resizebox{0.93\textwidth}{!}{
\includegraphics{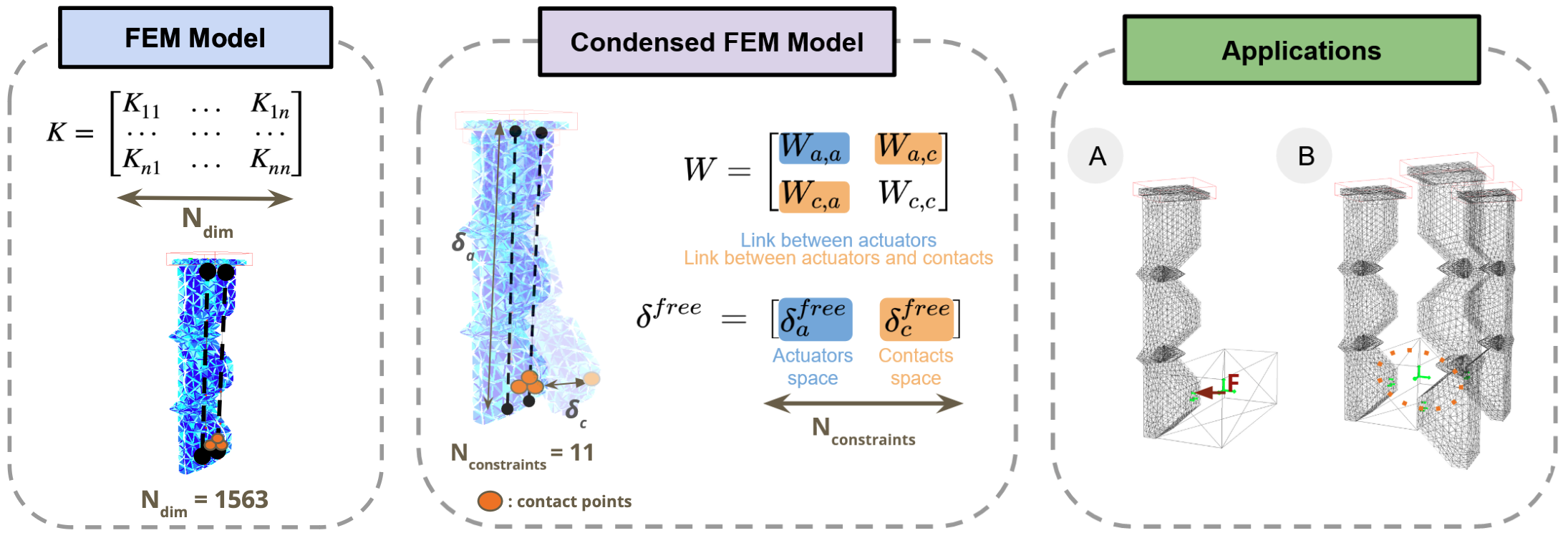}
}
\caption{Illustration of the condensed FEM model for a Soft Finger with contacts. The Soft Finger is actuated by 2 cables and has 3 fixed contact points located at the tip. The FEM model of the robot (left) is used to learn a condensed model (middle). The condensed model is based on the projection of the compliance matrix into the constraint space, resulting in drastically reducing the number of characteristic dimensions describing a robot state. The learned model is a data-driven supervised MLP model that predicts the projected compliance $W$ and the characteristic constraint distances $\delta^{free}$, based on the active constraint state (actuation, contact) and the initial values $W_0$ and $\delta_0^{free}$. On the right, two control scenarios are considered: A) a Soft Finger robot pushing a button; and B) a Soft Gripper robot manipulating a cube. The cube effector is represented by a green frame.
}
\label{fig:robots_contact}
\end{figure*}

In the FEM framework, computing $W$ is computationally expensive, as it requires the multiplication and inversion of several large matrices, which have to be recalculated at every simulation step. This provides a strong motivation to learn $W$ instead of computing it from mechanical simulation. Learning these mechanical quantities, rather than using unsupervised methods, enables the creation of interpretable mechanical robot representations usable for simulating, controlling, and even designing robots. The method is illustrated in 
Figure~\ref{fig:robots_contact}.

To build the condensed FEM model, we consider the following assumptions:
\begin{itemize}
    \item The robot's behavior is described using quasi-static mechanical equations. This assumption is valid for robots that either move slowly or have negligible mass. When this assumption no longer holds, dynamic effects arise that are not captured by the model. This assumption applies to many robots (catheters, trunks, etc.)~\cite{Coevoet2019, catheter2020}. However, for other applications, such as high-frequency pick-and-place~\cite{ZughaibiPickandPlace2021} or locomotors not attached to a fixed base~\cite{Coevoet2019}, dynamic effects often have to be considered. In this article, we focus on fixed-base robots with low speeds.
    \item The location of the contact points on the robot is known in advance, although these points are not necessarily active throughout the simulation. This assumption means that once the model is learned, it cannot account for contacts at other locations on the robot. This may limit certain applications. For instance, in some scenarios, such as dexterous manipulation~\cite{Coevoet2019} or catheter locomotion in minimally invasive surgery~\cite{catheter2020}, the position and number of contact points are not known in advance. 
    \item Only actuation and contact forces modify the state of the robot. This implies that even though $W$ directly depends on the global FEM state, knowing the states of both the actuators and contacts is sufficient to evaluate $W$ in the quasi-static case.
    \item The material models are assumed to be not highly non-linear, and non-linear phenomena such as buckling are not considered. Considering that the active constraints (including actuation and contact) are known, the robot can only reach a single deterministic configuration for a given set of constraints.
    \item The data space is dense enough to represent the robot's quasi-static state across many configurations. This assumption leads to significant sampling, which can be difficult to achieve when there are many different actuators or contacts. For example, sampling with a uniform grid strategy has exponential complexity regarding the number of actuators and contacts.
\end{itemize}

The learning problem can be formulated as follows. Let $\delta_{a}$ and $\delta_{c}$ represent the effective states reached by both actuation and contact constraints when a specific active constraint state is applied, and let $W$ and $\delta^{\text{free}}$ be the corresponding compliance matrix and free displacement of the robot projected in the constraint space. The goal is to reconstruct $(W, \delta^{\text{free}})$ from initial values $(W_0, \delta^{\text{free}}_0)$ for forces $\lambda_{a}=\bold{0}$, $\lambda_{c}=\bold{0}$ and current active constraint state $\delta_{a}$, $\delta_{c}$. This leads to learning a function $F$ such that:

\begin{equation}
\widetilde W, \widetilde \delta^{\text{free}} = F(\delta_{a}, \delta_{c}, W_0, \delta^{\text{free}}_0)
\label{global_function}
\end{equation}

$W_0$ is a good initial compliance matrix, as it already captures the mechanical coupling between constraints, and the compliance matrix varies smoothly from this initial position in the robot's workspace. 

\subsection{Learning Model}
\label{subsec:learningmodel}

The compliance matrix $W$, introduced in Equation~\ref{eq:schur}, is a symmetric matrix representing the compliance in the space of defined constraints: actuator, effector, and contact.  It is constructed from the mechanical matrices introduced in Equation~\ref{eq:schur}:

\begin{equation}
W = 
\begin{bmatrix}
W_{ee} & W_{ea} & W_{ec}\\
W_{ae} & W_{aa} & W_{ac}\\
W_{ce} & W_{ca} & W_{cc}\\
\end{bmatrix}
\label{eq:W}
\end{equation}

The magnitude of both $W$ and $\delta^{\text{free}}$ depends on both the type of constraints and the measurement units chosen for the simulation. To prevent vanishing gradients during the learning process, both the input and output data are element-wise standardized. This standardization is computed once using all the training sets.

Since the matrix is symmetric, the networks are trained in a supervised manner by minimizing the reconstruction loss: 
\begin{equation}
L = \sum_{\delta_a, \delta_{c}} [||\widetilde W^{\text{tri}} - W^{\text{tri}}||^2 +  ||\widetilde  \delta^{\text{free}} - \delta^{\text{free}}||^2]
\label{eq:Loss}
\end{equation}
where $W^{\text{tri}}$ is the upper triangular part of the matrix $W$, $\widetilde W^{\text{tri}}$ and $\widetilde  \delta^{\text{free}}$ are the predicted values of $W^{\text{tri}}$ and $\delta^{\text{free}}$ provided by the network for the actuation displacement $\delta_a$ and contact displacement $\delta_c$. Since all data are standardized, the computed loss functions obtained from different training data are somewhat comparable.

Well-known neural network architectures such as the MultiLayer Perceptron (MLP)~\cite{Haykin1994MLP} are sufficient for learning a mechanical representation to solve control and design tasks. This type of network is chosen for its simple implementation: it has few weights and produces satisfactory results quickly. In our implementation, the input consists of the raw data vectorized and concatenated to form a single vector, while the output is the vectorized pair $(\widetilde W^{\text{tri}}, \widetilde \delta^{\text{free}})$. Rectified Linear Unit (ReLu) activation functions are used.

Since the framework is based on supervised learning, pre-computation steps are first performed, where data are collected and the learning model is trained. The data $(\delta_{a}, \delta_{c}, W^{\text{tri}}, \delta^{\text{free}})$ are obtained through simulation and stored in a database. Each data point corresponds to a robot configuration after applying an active constraint displacement ($\delta_{a}$, $\delta_{c}$). The possible values of both $\delta_{a}$ and $\delta_{c}$ are bounded within continuous intervals. 

Sampling the active constraint space is done using Scrambled Halton (SH) sampling algorithm~\cite{{Mascagni2004}} to build the training dataset and random search for the test dataset. Compared to homogeneous grid search strategies, as shown in our previous work~\cite{Menager2023}, SH sampling allows for more uniform and efficient coverage of high dimensional space, providing better scalability for robots with numerous actuators and contacts. For example, while 15000 configurations sampled with a grid search strategy were used for the Soft Finger robot without contact, we reduced the number of required samples to 6500 using the SH sampling algorithm (see section~\ref{seq:designOptim}). Reducing the dataset by 60\% does not compromise the accuracy of the resulting matrices.  Furthermore, the number of sampled configurations is comparable to those required by other methods. For instance, even though the two reduced models are designed for different purposes - full spatial configuration reconstruction versus control of specific points -, the POD-based model presented in~\cite{Goury2018} requires 10000 samples to create a reduced model of a cable-driven parallel robot, whereas the learned condensed FEM model relies on only 6561 samples for the same robot with a grid search strategy in~\cite{Menager2023}. While we do not optimize the number of samples in this article, we note that there is potential for improvement in this area.

The number of points of the test set is set to $25\%$ of the number of points of the training set. Once the data is acquired, the learning process involves training the network to predict the quantities $(W^{\text{tri}}, \delta^{\text{free}})$ from the input $(\delta_{a}, \delta_{c}, W^{\text{tri}}_0, \delta_0^{\text{free}})$ using MLP predictions and Equation~\ref{eq:Loss}. Unless stated otherwise, the network used in the experiments presented in this paper is a fully connected MLP with 3 hidden layers of 450 nodes each. The learning rate is initialized to $10^{-3}$ and an adaptive scheduler is consistently used in all experiments to automatically reduce the learning rate when the optimizer encounters a local optimum. 

For all the learning experiments, training stops after a maximum of 50000 epochs. We retain the model from the epoch that achieves the lowest loss on the test set. A discrete hyperparameter search is performed to determine the best set of hyperparameters. Various combinations are tested, and models are compared using our training strategy on the same dataset. The hyperparameters include the number of hidden layers, ranging from 3 to 5, and the number of nodes per hidden layer, ranging from 400 to 900.

\subsection{Learned Condensed FEM Model for Control Application}
\label{ControlScheme}

Once learned, the condensed FEM model can be used in a control loop. We assume that $W^{\text{tri}}_0$ and $\delta_0^{\text{free}}$ are known, while $\delta_{a}$ and $\delta_{c}$ are obtained through measurements on the robot. The trained network is then used to predict $(\widetilde W^{\text{tri}}, \widetilde \delta^{\text{free}}) = F(\delta_{a}, \delta_{c}, W^{\text{tri}}_0, \delta_0^{\text{free}})$, where $\widetilde W$ is reconstructed from $\widetilde W^{\text{tri}}$.  

Even if the condensed FEM model is trained on a predefined set of fixed goals $x_{\text{goal}}^{\text{train}}$, one for each of the effectors according to $\widetilde \delta_{e}^{\text{free}} = x_{\text{effector}} - x_{\text{goal}}^{\text{train}}$, it can be easily adapted to other goals $x_{\text{goal}}$ during the control phase using the following formula:

\begin{equation}
     \delta_{e}^{\text{free}} = \widetilde \delta_{e}^{\text{free}} + x_{\text{goal}}^{\text{train}} - x_{\text{goal}}
\label{eq:generalizeGoal}
\end{equation}

The learned quantities are used in an inverse optimization problem, such as the one in Equation~\ref{eq:QPActuationContacts}. This type of problem is solved using a quadratic optimization solver. The goal is to compute the actuation force vector $\lambda_{a}$ needed to control the robot. In our work, we use the C++ library Prox-QP \cite{Bambade2022} to implement the solver. Once the actuation force $\lambda_{a}$ is applied on the robot, the respective new actuation and contact states $\delta_{a}$ and $\delta_{c}$ are recovered. At this stage, corrective motion can be applied to improve performance, stability, and precision. In particular, it helps to address the two main sources of errors: those caused by the learning process and the simulation-to-reality gap. The control loop then repeats with a new prediction until convergence, i.e. until an accuracy or stability criterion is met.

In section~\ref{sec:controlWithContact}, this pipeline is applied to simulated robots. Since the simulation-to-reality transfer for the robots used in our experiments has already been studied in previous works, the simulation can be considered as ground truth for evaluating our algorithms. For a FEM model, this transfer is possible when the mechanical and geometric parameters of the simulated model are well calibrated, the mesh discretization is sufficiently precise, and when the boundary conditions on forces and contacts are respected. Constraints states can be directly obtained from the simulation results. In section~\ref{sec:embeddedControl}, this workflow is applied to a physical prototype, where constraint states are retrieved from a set of sensors, thus avoiding the need for FEM simulation of the robot within the control loop. 

\section{Soft Robot Control using a Condensed FEM Model with Contacts}
\label{sec:controlWithContact}

In this section, we introduce contact modeling using the condensed FEM model, mostly for manipulation control.

\subsection{Soft Gripper Robot with Contacts}

To evaluate the condensed learned model with contact, we consider a Soft Finger robot model, validated in \cite{navarro2020model}. This robot comprises three segments connected by accordion-shaped joints and actuated by two cables. This robot is simulated in SOFA \cite{SOFA} using a mesh of 1557 tetrahedrons. Since the simulation-to-reality transfer has already been studied in the work mentioned above, the simulation results are considered as the ground truth for our study.

First, the Soft Finger is used to press a cube-shaped button, with the button's resistance modeled as a constant horizontal force. Then, three identical Soft Fingers are combined to create a gripper for manipulating an object. This example demonstrates that a single condensed FEM model, learned for one Soft Finger robot, can be applied in an inverse optimization scheme involving multiple Soft Fingers. We made several simplifying assumptions: (i) the gripped object is always rigid, (ii) the contact locations on the actuated flexible finger are fixed and placed on the surface of the last phalanx, and (iii) the contact points are always active and frictionless. The condensed FEM model and the two application cases are illustrated in Figure~\ref{fig:robots_contact}.

When generating the dataset for learning the condensed FEM model for both scenarios, the sampling space is directly proportional to the number of actuation and contact constraints. One challenge encountered is that exploring the constraint space has exponential complexity as the number of constraints increases. In the case of the Soft Finger, there are 2 constraints related to actuation and 9 constraints related to contacts. 
Moreover, using a box sampling on both the actuation and contact constraint spaces would result in many robot states irrelevant to our applications. 
For this reason, in both cases, the dataset is generated by sampling in the cube effector displacement space and using an inverse model to find the corresponding robot configuration. This means that the position of the cube is sampled first, and then an inverse model is used to determine the active constraints required to reach it. As a result, the sampling space size is reduced from 11 constraints to 3. Since the two scenarios involve different cube configurations, two separate training sets are generated, each time for a single Soft Finger. 
The respective final test losses for training the condensed FEM model for the Soft Finger and one Soft Finger of the Soft Gripper over 50000 training episodes are 1.67e-6 and 1.24e-3. 

\subsection{Inverse Problem for Soft Finger/Object Interaction}

Let $X_{\text{cube}} \in \mathbb{R}^6$ represent the position of the cube, including both translation and rotation. Let $^{\text{prev}}$ denote quantities calculated at the previous time step in an iterative process. In a manipulation task, our goal is to control the translation of the cube, which is equivalent to minimizing the following quantity:

\begin{equation}
    \underset{\lambda_a}{\text{min}} || X_{\text{cube},\text{trans}} - X_{\text{cube}}^{\text{goal}}||^2
    \label{eq:optcontact}
\end{equation}

where $X_{\text{cube}}^{\text{goal}}$ is the translation component of the goal position, $X_{\text{cube}, \text{trans}}$ is the translation component of $X$ and $||.||$ is the Euclidean norm. To solve this problem, $X_{\text{cube}}$ has to be expressed as a function of the actuation force $\lambda_a$. 
We use condensed mechanics and cube kinematics to obtain this function. 

\subsubsection{Kinematics of Points Placed on a Rigid Object} \hfill

Based on the assumption that the contact points locations on both the Soft Finger and the object are shared, it results that $\lambda_c = \lambda_{c, \text{finger}} = - \lambda_{c, \text{cube}}$ and $x_{c, \text{finger}} = x_{c, \text{cube}}$, where $x_{c, \text{finger}}$ and $x_{c, \text{cube}}$ are the Euclidean positions of the contact points between the Soft Finger and the cube. 

For an infinitesimal displacement of $X_{\text{cube}}$, the positions of $X_{\text{cube}}$ and $x_{c, \text{cube}}$ are related as:
\begin{equation}
    x_{c, \text{cube}} - x_{c, \text{cube}}^{\text{prev}} =   J_c (X_{\text{cube}} - X^{\text{prev}}_{\text{cube}}) 
    \label{eq:linkxcXcube}
\end{equation}
where  $J_c \in \mathbb{R}^{N \times 6}$ is the Jacobian matrix that links the kinematics of the manipulated object to the contact points, and $N$ is the number of contact points. 

Using this Jacobian matrix, the resulting contact force on the cube is expressed as $f_{\text{cube}} = J_c^{T} \lambda_c$.\\

\subsubsection{Condensed mechanics applied on a Soft Manipulator} \hfill

Given the contact forces $f_{\text{cube}}$ imposed on the cube, and multiplying by $D = J_c^{T} W_{cc}^{-1}$, the following relation is obtained: 

\begin{equation}
\begin{aligned}
     D x_{c, \text{cube}}^{\text{prev}} + &D J_c (X_{\text{cube}} - X^{\text{prev}}_{\text{cube}}) = \\
     &D \delta_{c}^{\text{free}} + D W_{cc} \lambda_c + D W_{ca}  \lambda _a 
     \label{eq:poslambdaalambac}
\end{aligned}
\end{equation}

By manipulating the equation, we obtain a function $X_{\text{cube}} =  \mathbb{A}\lambda _a + \mathbb{B}$, since $D W_{cc} \lambda_c = f_{\text{cube}}$ is known. The optimization objective from Equation~\ref{eq:optcontact} is now well-posed.

The Soft Finger robot and the Soft Gripper robot differ in the way the force $f_{\text{cube}}$ is computed. From the new values of $X_{\text{cube}}$ and $\lambda_a$, it is then possible to compute a new value of $\lambda_c$ using Equation~\ref{eq:poslambdaalambac}.

\subsection{Application: Soft Finger pushing a Press-button.}

\begin{figure}[!ht]
\centering
\resizebox{0.5\textwidth}{!}{
\includegraphics{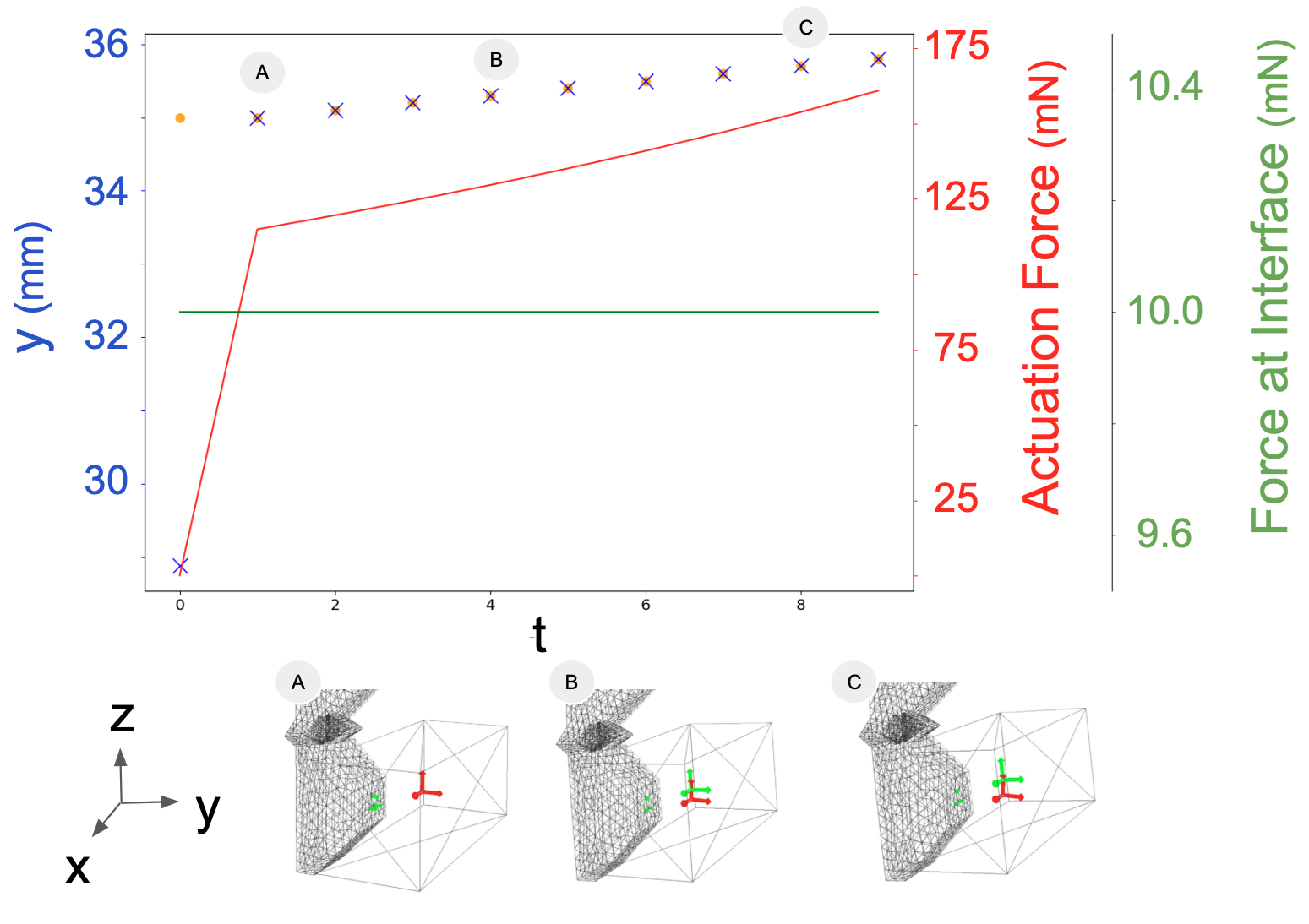}
}
\caption{Control results for the Soft Finger robot pressing a button. The reached horizontal position (mm) of the cube is compared on a trajectory of 9 successive goals (orange dots) when using learned mechanical matrices (blue cross). The contact force exerted at the interface between the Soft Finger and the button as well as the sum of actuation forces exerted on the two cables during the trajectory are also displayed. 3 different positions of the Soft Finger robot met during this trajectory are shown as examples. In this representation, the effector is a green frame, and the goal is a red frame.}
\label{fig:control_1fingercontact}
\end{figure}

We aim to control the position of a cube through its contact with the soft finger.
A known constant horizontal force $F$ opposes the direction of the applied bending movement. The contact force $f_{\text{cube}}$ can be deduced from $F$. 
A trajectory of positions for the center of the cube $X_{\text{cube}}$, following a horizontal line, is given as the target. Since the main direction of deformation of the Soft Finger is in its bending direction, only a single degree of freedom of the cube is controlled. 
The results are shown in Figure~\ref{fig:control_1fingercontact}.

In this application, the open-loop control of the soft robot using the condensed FEM model demonstrates a similar level of precision to that of simulation-based control.

\subsection{Application: Soft Gripper Manipulating a Cube.}

\subsubsection{Mechanical Coupling Using Contacts for Manipulation}

The mechanical matrices of the Soft Gripper robot are built from the mechanical matrices computed for three individual Soft Finger. Unlike the scenario where the Soft Finger pushes a button, the applied forces on the cube are not known. 

After solving the same problem as for a single Soft Finger, Equation~\ref{eq:poslambdaalambac} enables the expression of the contact forces $\lambda_c$ based on the new values of the actuation forces $\lambda_a$ and the 6D position of the cube $X_{\text{cube}}$. The corresponding algorithm is presented in Algorithm~\ref{alg:contact}.

\begin{algorithm}
\caption{Control algorithm for object manipulation scenario.}
\label{alg:contact}
\begin{algorithmic} 
\ENSURE Start from $\lambda_c = 0$.
\FOR{each time step}
\STATE 1. Compute $\lambda_a$ from Equation~\ref{eq:optcontact} using $ X_{\text{cube}} =  \mathbb{A}\lambda _a + \mathbb{B}$ where $\mathbb{B}$ is computed using Equation~\ref{eq:poslambdaalambac} and the previous evaluation of $\lambda_c$.
\STATE 2. Evaluate the new value of $\lambda_c$ using Equation~\ref{eq:schur} and the new value of $\lambda _a$.
\ENDFOR
\end{algorithmic}
\end{algorithm}

In this algorithm, both $\lambda_a$ and $\lambda_c$ are updated once per time step using their previous values. Experiments showed that this scheme is sufficient to control the robot. Another valid approach would be to use an internal iterative procedure to gradually update $\lambda_a$ and $\lambda_c$ until convergence at each time step.
With the design choices for the Soft Gripper robot, only small movements of the cube are possible around its resting position. This is because the initial configuration of the Soft Fingers in contact with the cube consists of fully relaxed cable actuation, limiting motion in the opposite direction of the Soft Fingers' bending. \newline

\subsubsection{Results}

\begin{figure}[!ht]
\centering
\resizebox{0.5\textwidth}{!}{
\includegraphics{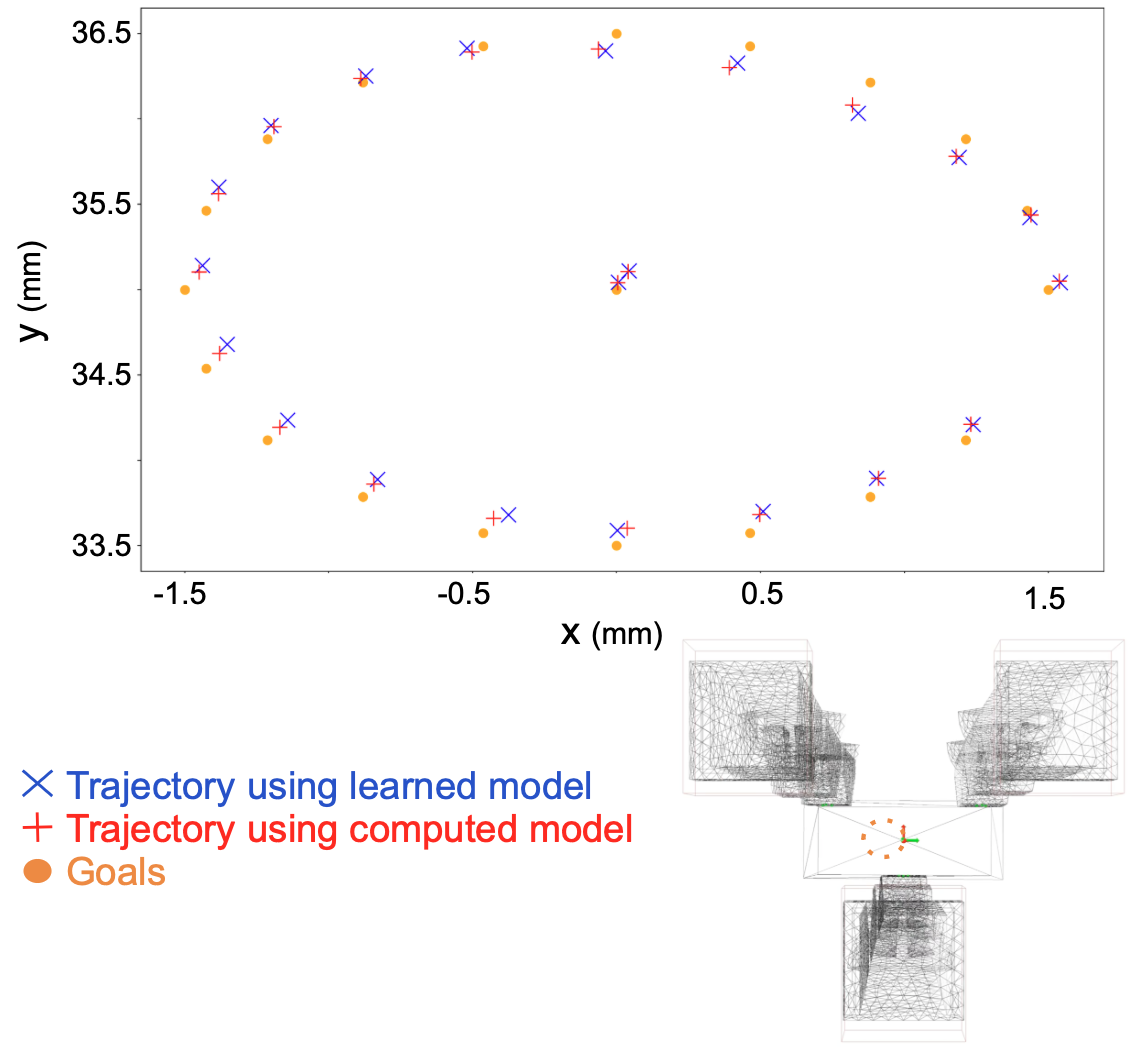}
}
\caption{Control results for the Soft Gripper robot. The positions of the manipulated cube (mm) are compared on a trajectory for 20 different goals (orange dots) when using learned mechanical matrices (blue cross) or mechanical matrices computed from the full simulation (red cross).}
\label{fig:control_grippercontact}
\end{figure}

For the experiment, a circular trajectory of the object's center is considered. The results are shown in Figure~\ref{fig:control_grippercontact}.

Along the considered trajectory, slight differences between the target and the reached effector positions are observed with both the simulation-based and learning-based controllers. This can be explained by the modeling of the manipulator, which assumes that the Soft Fingers' contact points remain fixed on the manipulated object. These errors could be reduced by using a closed-loop controller.

Additionally, small differences between the reached positions of the simulation-based and learning-based controllers are observed. Compared to the case of the Soft Finger pushing a button, the training results are less accurate. This is due to the increased complexity of the dataset, as various forces can be applied to the cube. 

Currently, the application considers only small deformations. However, the results demonstrate that contact information can be integrated into the learned condensed FEM model, which is promising. Future works may focus on cases involving larger deformation and collision detection/resolution to eliminate the assumption of fixed contact points. This could pave the way for locomotion applications.

\section{Embedded Control of Soft Robot Using A Condensed FEM Model}
\label{sec:embeddedControl}

In this section, we demonstrate how the condensed FEM model can be used to control a physical pneumatic robot's ongoing large displacements without relying on online FEM simulation. This approach leverages both the small scale and the fast prediction speed of the neural network, making it suitable for applications in embedded control.

\subsection{Pneumatically Actuated Continuum Robot.}

\begin{figure*}[!ht]
\centering
\resizebox{0.8\textwidth}{!}{
\includegraphics{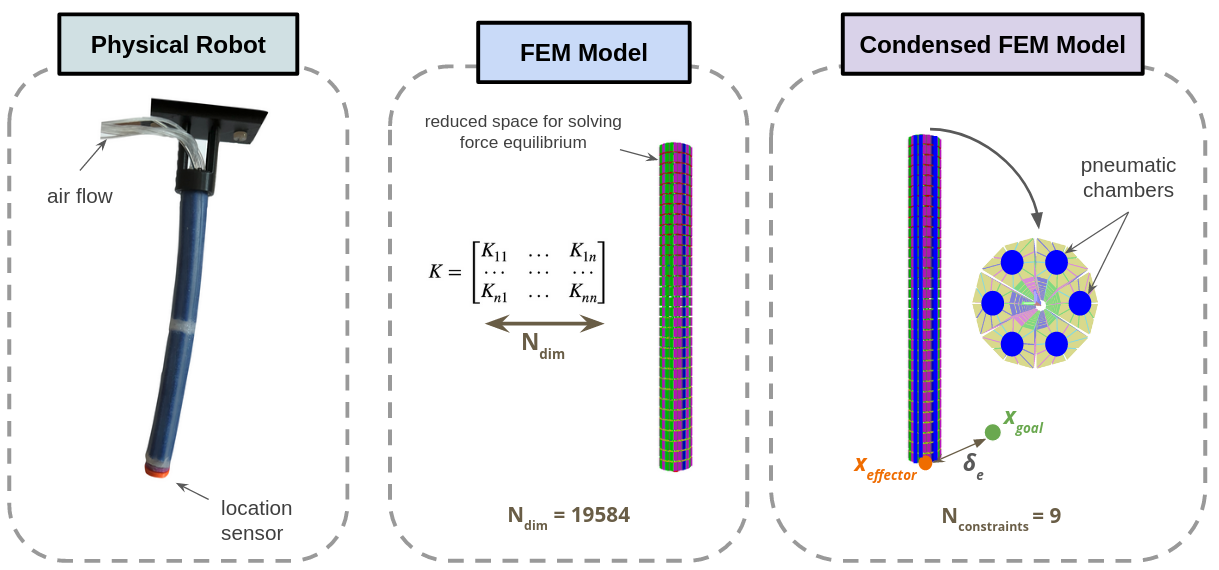}
}
\caption{Illustration of the physical prototype and the condensed FEM model for the Stiff-Flop robot. The robot is actuated by 6 pneumatic cavities represented by blue cylinders. A location sensor is located at its tip. The reduced FEM-based modeling from~\cite{ChaillouRobosoft2023} is also displayed.}
\label{fig:robot_embedded}
\end{figure*}

The considered soft robot is a pneumatic trunk robot initially introduced in~\cite{abidi_highly_2018}, called the Stiff-Flop robot. It is a highly flexible, pneumatically driven soft robot composed of two modules, each containing three fully fiber-reinforced chamber pairs. This robot was developed for medical applications, specifically as an in-vivo cancer diagnostic tool during minimally invasive interventions. The physical robot and the associated simulation are shown in Figure~\ref{fig:robot_embedded}.

\begin{figure}[!ht]
\centering
\resizebox{0.4\textwidth}{!}{
\includegraphics{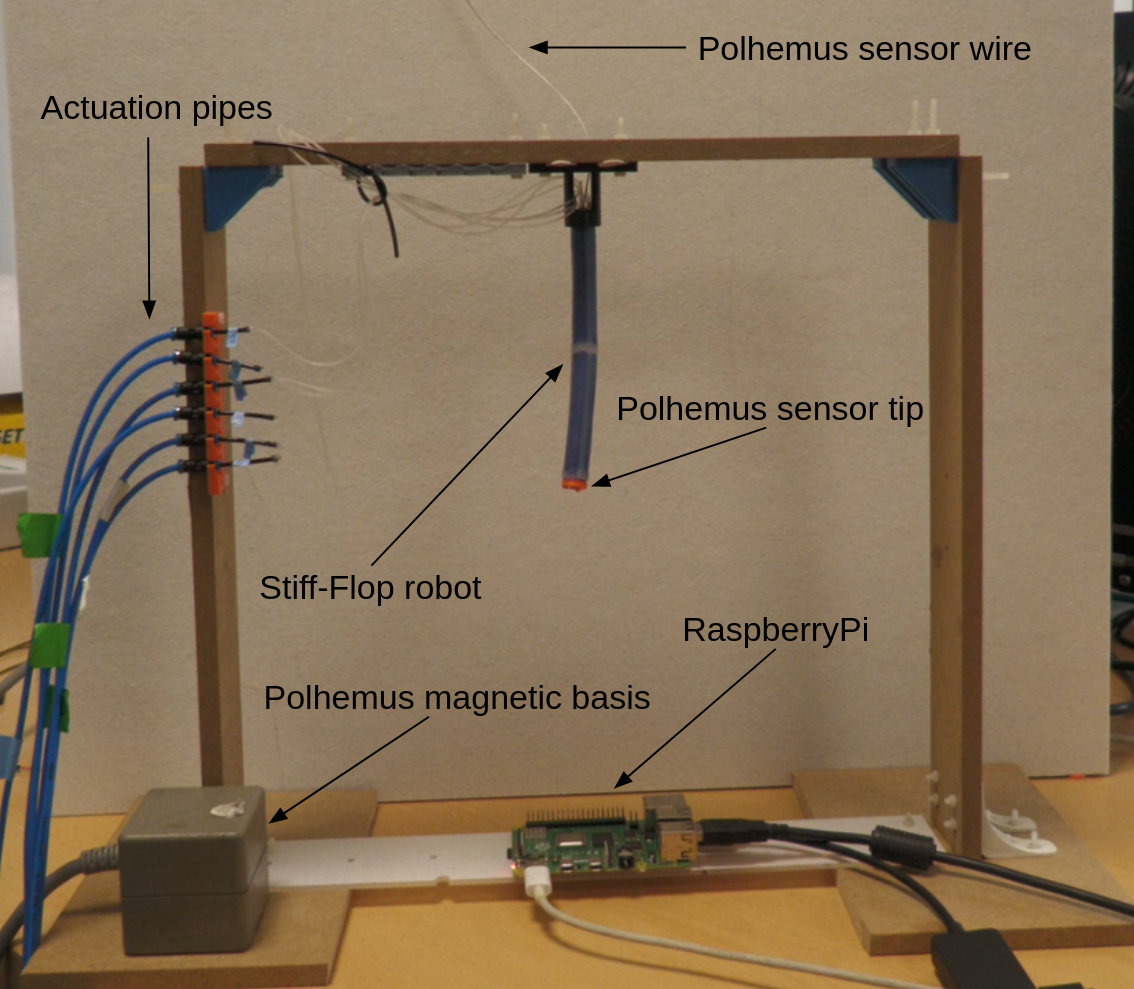}
}
\caption{Picture of the setup with the Raspberry Pi.}
\label{fig:Raspberry_Setup}
\end{figure}

\textbf{Physical prototype:} The chamber pressure is regulated and monitored by solenoid valves (Rexroth ED02 R414002399). These pressure regulators are voltage-actuated, and the proportional voltage is generated using a dedicated voltage output board (PhidgetAnalog 4-Output 1000$\_$0b). Pressurized air is supplied to the regulators by a compressor (Fengda FD-186). An electromagnetic tracking system (Polhemus Liberty) monitors the end-effector position and orientation. The system may be connected either to a computer or a Raspberry Pi, synchronized and linked with the learned model via Python scripts. An illustration of the hardware setup with a Raspberry Pi is provided in Figure~\ref{fig:Raspberry_Setup}.

\textbf{Simulation:} The Stiff-Flop robot is built using a beam-based reduced finite element model (BBRFEM) as described in~\cite{ChaillouRobosoft2023}. In contrast to~\cite{ChaillouRobosoft2023}, the sensor stiffness cannot be neglected in our physical prototype. Consequently, we adjusted the stiffness of the modules, resulting in a Young Modulus of 150kPa for the base module and 90kPa for the top module.

\textbf{Learned condensed FEM model:} The condensed FEM model is learned based on the BBRFEM rather than the full-FEM simulation to accelerate the sampling phase and to demonstrate the ability of the condensed FEM model to further decrease computation cost, even for an already optimized simulation. Data are collected in the space of the volume change of the cavities. The dataset consists of $30000$ robot configuration. The final test loss for training the condensed FEM model for this soft robot over 50000 training episodes is $6.05e-5$.

\subsection{Control Using the Condensed FEM Model and Minimal Sensor Feedback}

A key assumption of the control loop presented in section~\ref{ControlScheme} is that both the actual displacements of the actuators and the actual positions of the effectors are observable.
In practice, this requires the systematic use of a set of specialized sensors, introducing an additional constraint on the physical prototype. This issue is addressed by the following framework, which targets both embedded open- and closed-loop control of a robot using absolute position sensor feedback.

For the Stiff-Flop robot, the system is pressure-controlled, meaning there is no direct access to the actuator's position. In our case, this would require a sensor for the volume of each cavity, which is not trivial to set up. Instead, the condensed FEM model is used as a state observer. The volume of the cavities is computed based on the actuation $\lambda_a$ and the projected mechanical matrices from $\delta_a(x) = W_{aa}^{\text{prev}} \lambda_a + \delta_a^{\text{free}, \text{prev}}$ where $^{prev}$ refers to the evaluation of the mechanical quantities in the previous iteration of the control loop. This strategy introduces a time-step delay in the predicted mechanical matrices used to compute the actuation displacement. This delay can be corrected using an appropriate corrector in a closed-loop setting.

The general embedded control loop framework is as described in Algorithm~\ref{alg:embeddedloop}. 

\begin{algorithm}
\caption{Embedded control loop with end-effector location feedback only.}
\label{alg:embeddedloop}
\begin{algorithmic} 
\ENSURE Access to $W_0$ and $\delta^{\text{free}}_0$ from the training dataset. Set initial actuation  $\lambda_a = \bold{0}$ and initial actuators displacement $\delta_a = \delta^{\text{free}}_{a,0}$.
\FOR{each time step}
\STATE 1. If considering closing the loop, recover the position of the effectors and compute the corrected goal position.
\STATE 2. Predict the mechanical quantities $\widetilde W$ and $\widetilde \delta^{\text{free}}$ according to Equation~\ref{global_function} without contact, using the trained neural network.
\STATE 3. Solve the optimization problem to find the force $\lambda_a$ according to Equation~\ref{eq:QPActuationContacts} without contact.
\STATE 4. Compute the new actuator state from mechanical matrices and forces as $\delta_a = W_{aa} \lambda_a + \delta^{\text{free}}_a$ for querying the learned condensed FEM model at the next iterations. 
\STATE 5. Apply the forces $\lambda_a$ on the robot.
\ENDFOR
\end{algorithmic}
\end{algorithm}

To close the loop, a measurement taken from the physical robot must be compared with the values predicted by the control scheme. In the proposed experiment, a position sensor is placed on the tip of the robot, and a PI controller is applied to the effector displacement (see Equation~\ref{eq:generalizeGoal}). The same proportional and integral coefficients are used for all the experiments, regardless of the control loop frequency.

\subsection{Experiments in a Virtual Environment}

To evaluate the use of the learned condensed model for inverse control of the Stiff-Flop robot, a 40mm radius circular trajectory around the robot's tip is considered. This trajectory consists of a succession of 200 goal positions to reach the circle from the initial rest position, followed by 600 successive goal positions describing the circle itself. 

The objective of the following experiments is to evaluate the introduced control scheme used with the learned condensed FEM model in an open loop. Initially, the experiments are conducted in a virtual environment to avoid artifacts associated with the simulation-to-reality transfer. The positions errors are compared using either the computed state of the actuation $\delta_a$ or the predicted state of the actuation $\delta_a = W_{aa}^{\text{prev}} \lambda_a +\delta^{\text{free}, \text{prev}}_a$, as explained in the previous section. 

In both cases, the trajectory is closely matched in open-loop, with a maximum error of approximately 1.8 mm using the computed state and 5 mm using the predicted state. These results show that the robot is capable of following the circular trajectory, even without measuring actuation states in the simulation, as in the second case. When using the predicted state, a constant shift, depending on the discretization step used for the integration, is nevertheless observable. Increasing the number of intermediate goals sampled along the circular trajectory reduces the error in the robot's state estimation between successive points. This shift can also be compensated by implementing a closed-loop control scheme based on minimal sensor feedback, as demonstrated in the following experiments on the physical prototype.

\begin{figure*}[!ht]
\centering
\resizebox{0.9\textwidth}{!}{
\includegraphics{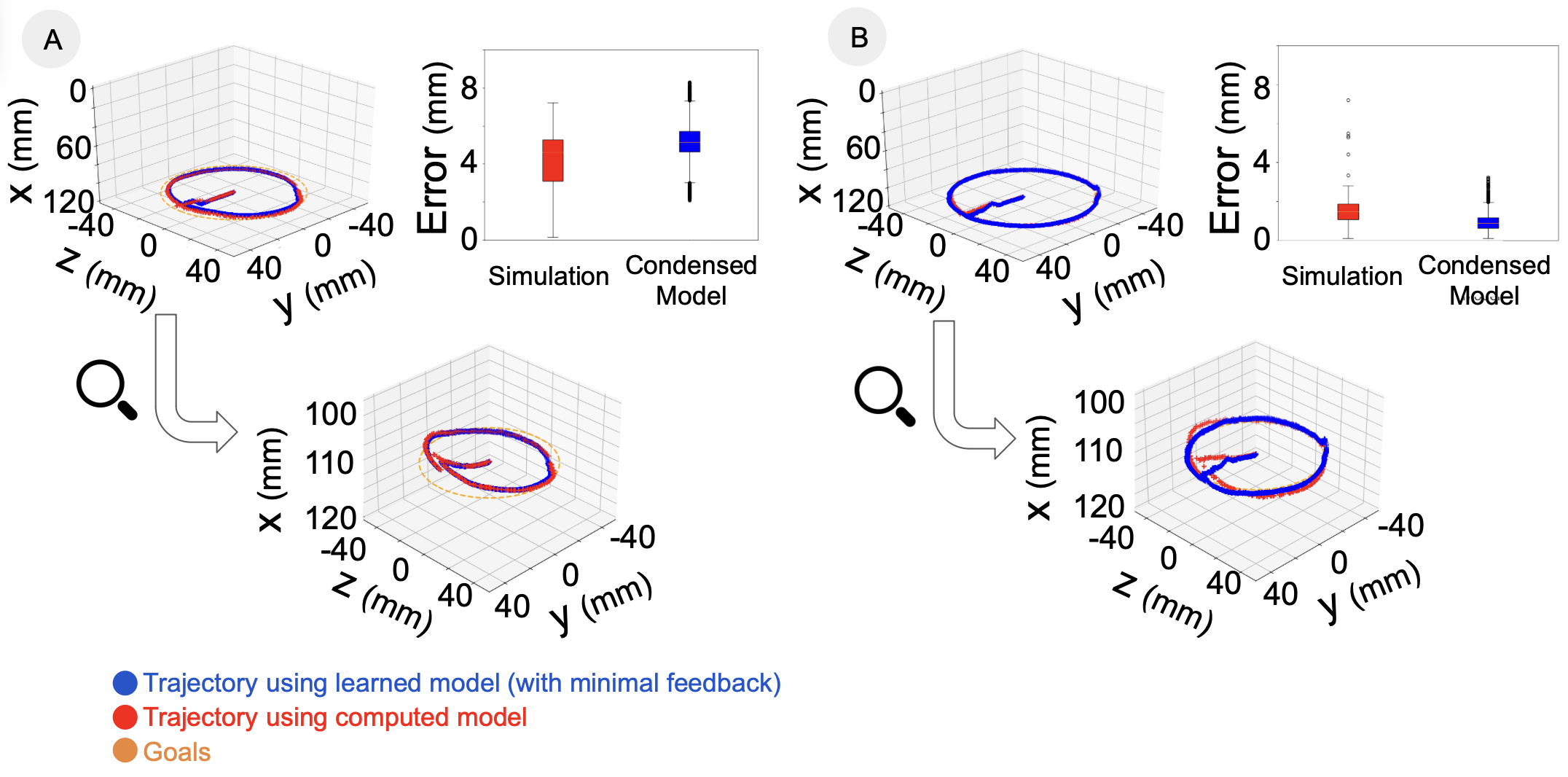}
}
\caption{ 
Trajectory and tracking errors were obtained for performing a circular trajectory (orange dots) using condensed FEM model-based control schemes with the physical prototype of the robot. Errors are computed as the absolute distance between the desired and the reached positions. Experiments are A) Comparison of simulation (red) based and condensed FEM model with minimal sensor feedback (blue) based open-loop control, B) Comparison of simulation (red) based and condensed FEM model with minimal sensor feedback (blue) based closed-loop control.}
\label{fig:StiffFlop_Traj_Results}
\end{figure*}

\subsection{Experiments with the Physical Prototype}

In addition to experiments conducted in simulation, the control is tested on the physical robot. Results are shown in Figure~\ref{fig:StiffFlop_Traj_Results}. Both open- and closed-loop scenarios follow the same 40mm circular trajectory. This demonstrates both the benefits and drawbacks of the learned condensed FEM model for positioning purposes on a physical prototype.

The computational cost reduction allows the learned model to run at 1 kHz. However, communication with the hardware setup limits the feedback loop at 90 Hz, primarily due to the Polhemus sensor. With the BBRFEM, the simulation runs at 5 Hz, so the communication system does not restrict the control loop frequency. 

In the rest of this section, we will detail and discuss the results obtained with the physical prototype in terms of open-loop and closed-loop control, the convergence speed of the closed-loop control, and embedded control. 

\paragraph{Open-Loop Control}

Since computing the robot's state using the learned condensed FEM model is significantly faster than using the BBRFEM, the actuation pressures are also applied quickly. This can lead to dynamic effects that are not accounted for in the quasi-static modeling used to learn the condensed FEM model. To reduce these dynamic effects, the number of intermediate goals sampled on the trajectory is increased. At each time step, the robot's inverse problem is solved as a quadratic problem (QP), and the actuation needed to move from one point to another is applied. By controlling both the position of the robot (by setting goals along the trajectory) and its speed (by sampling the trajectory according to the controller time step), the speed and variation in speed remain compatible with quasi-static assumptions. The greater the number of points on the circle, the less dynamic effects impact the model's performance. The quasi-static assumption works better when the number of points on the trajectory is increased for a given time budget. This criterion is used to compare the temporal performance of the models over several goals, as well as the accuracy they achieve within a given time budget. In practice, we assume the same calculation time budget for both models to achieve a circular trajectory, which is set to 80 seconds.

Regarding the computational performance of the model, the goal is to determine how many points can be considered within the allocated time. Since the model has to be computed for each point to be reached, this value represents the average calculation time for both models. Over 80 seconds of calculation, 3300 points can be reached with the learned condensed model, while only 500 points can be reached with the BBRFEM. On average, the open-loop control using the learned condensed FEM model is 7 times faster than the BBRFEM.
 
Regarding model accuracy under time budget constraint, results are shown in Figure~\ref{fig:StiffFlop_Traj_Results}.A. It demonstrates the ability to transfer to a physical prototype using the learned condensed FEM model with minimal sensor feedback. The trajectory errors when using the learned condensed FEM model are of the same order of magnitude as when using the BBRFEM. The small difference in mean error could be explained by the use of predicted actuation states from the previous mechanical state, leading to a one-time step delay between the mechanical state values computed by the two methods.

Thus, using the condensed FEM model does not increase the speed of completing the circular trajectory under strict accuracy constraints for the Stiff-Flop robot. Indeed, increasing the speed directly increases the positioning error due to the dynamic effect. However, it is possible to consider more points in the trajectory for the same time budget with the learned condensed model. This enables finer control of the robot's speed through a denser sampling along the trajectory and faster compensation of errors in a closed-loop setting. This is demonstrated in the following experiment.

\paragraph{Closed Loop Control}

The closed-loop trajectory is evaluated with the same iteration distribution and time budget as in the open-loop experiment. The proportional and integral error weights are set to $K_P = 0.01 $ and $K_I = 0.002 $. The choice of the control approach is not the focus of this paper. The objective is to demonstrate the utility of the condensed model in a closed-loop control scenario, without aiming to design the optimal closed-loop control for this task. More efficient control schemes could be developed.

Results are shown in Figure~\ref{fig:StiffFlop_Traj_Results}.B. The improvement in positioning accuracy is significant when using the closed-loop control based on the learned condensed FEM model. This accuracy improvement is primarily due to the PI correction. The correction frequency is much higher with the learned model, allowing the system to better compensate for both modeling and control errors compared to the closed-loop using the BBRFEM. For the same time budget, the learned condensed FEM model allows for more iterations, enabling more successive corrections to achieve more precise positions.  
The high computation frequency of the learned condensed FEM model effectively compensates for inaccuracies in closed-loop control applications. 

\subsubsection{Convergence Speed of the Closed Loop Control}

In this experiment, the convergence speed of the closed loop is evaluated by assessing the ability to reach a point in space with a given precision, i.e. when the error relative to the target is below a certain threshold. Practically, the error threshold is met when the error remains below $\epsilon=0.5$mm for at least 2.4 seconds, corresponding to approximately 10 simulation steps of the BBRFEM.

\begin{figure}[!ht]
\centering
\resizebox{0.5\textwidth}{!}{
\includegraphics{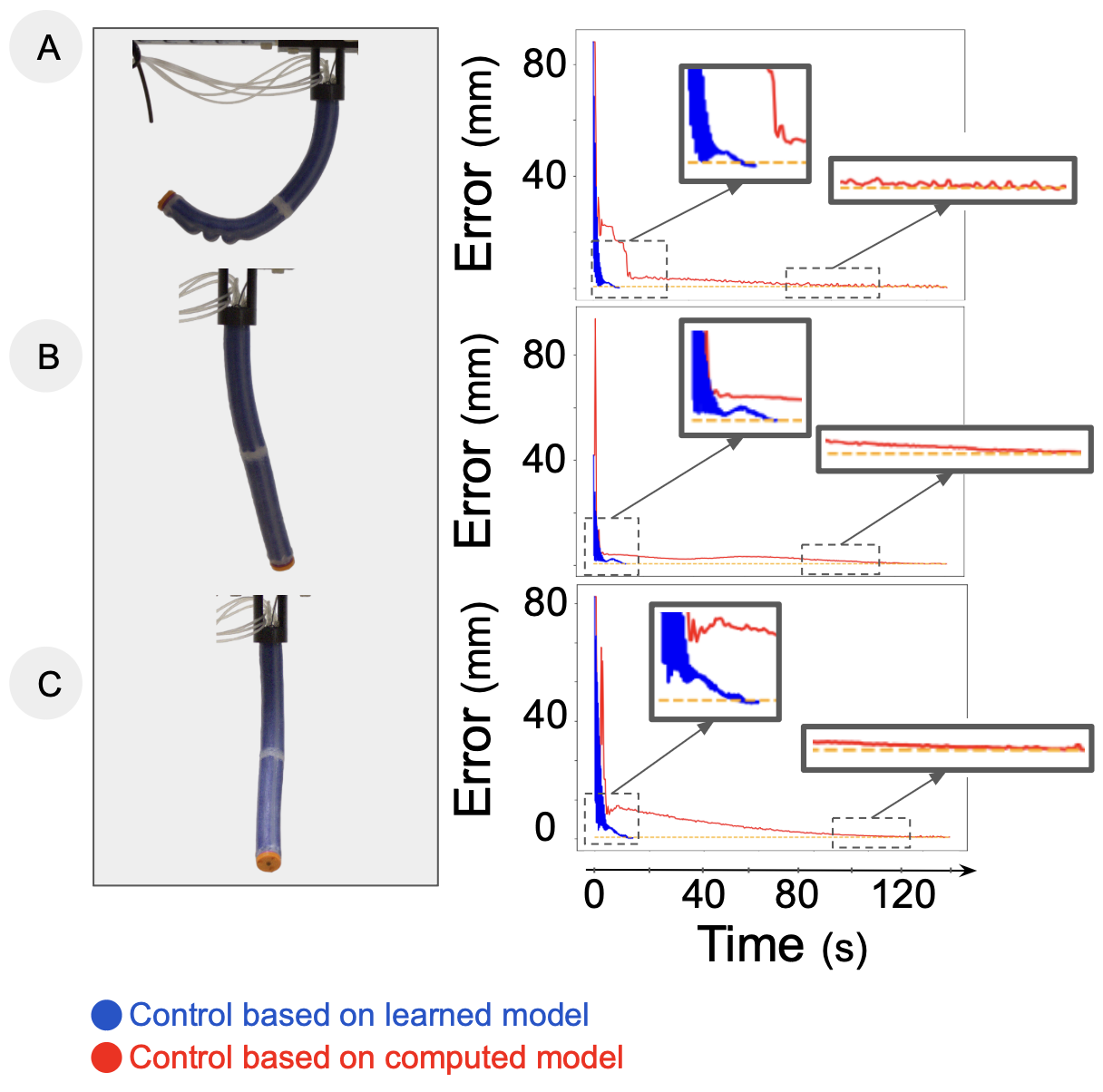}
}
\caption{Convergence histories obtained in a closed loop setting for reaching a point, and pictures of the associated reached state by the physical prototype. Log-scale error histories are compared respectively for simulation-based (red) and condensed FEM-based (blue) closed-loop control. The aimed error is 0.5mm to the target point (orange dot line). Considered target points are A) [57, -5, -70], B) [106, 26, 32.5], and C) [92.5, -58.5, 9].}
\label{fig:CLPointsHisto}
\end{figure}

The results are shown in Figure~\ref{fig:CLPointsHisto}. The error threshold is reached in over 100 seconds with the BBRFEM, compared to around 10 seconds for the learned condensed model. Thanks to its high correction speed, the closed-loop control based on the learned condensed FEM model reaches the desired position faster than with the BBRFEM control loop. This demonstrates that a PID corrector can iteratively correct errors more quickly with the learned condensed model, as its predictions are faster than those of the BBRFEM, and the error between the two models remains small.

This speed improvement highlights the utility of the learned condensed FEM model for accurate positioning control. Additionally, as shown in the supplementary video, this high-frequency correction enables the system to react faster to external disturbances, such as contact, without explicitly modeling them. If the closed-loop system converges, it necessarily drives the error to zero due to the integrator, enabling it to reject constant disturbances. The robustness of this type of control is explored in~\cite{ControlAlex}, concluding that the key factors are the direction and amplitude of the required actuation, specifically the quality and uncertainty of the Jacobian $W$. In the considered case, the disturbance is not strong enough to alter the structure or the principal directions determined by this matrix, thus ensuring convergence. Depending on the task, this feature could be an asset to ensure a good balance between precision and speed along a given trajectory.

\subsubsection{Embedded Control}

The control loops developed using the learned condensed FEM model have low computation and memory resource consumption. The memory usage of the main controller loop is estimated at 660 MiB on average. The entire algorithm has been embedded into a Raspberry Pi 4 and tested on different scenarios, yielding comparable results. This demonstrates the feasibility of using the embedded control loop based on the condensed model on microprocessors. The positionning results are identical to those presented in Figure~\ref{fig:StiffFlop_Traj_Results}. The main difference lies in the calculation time, which is longer on a Raspberry PI 4 than on a computer with a high-performance processor. This implementation highlights the clear advantage of using the learned condensed FEM model for embedded robot control, as it can run with good performances on light hardware. This could be particularly beneficial for control in autonomous robot applications.

\section{Condensed FEM Modeling Extended to Calibration And Design}
\label{seq:designOptim}

In traditional design optimization loops, designs are evaluated in simulation to iteratively update their parameters. The total computation time is directly related to how long it takes to simulate each design. Our model learns mechanical matrices instead of fitness functions and incorporates controller variables across the entire workspace. Specifically, the mechanical quantities $W$ and $\delta^{\text{free}}$ represent the connection between the robot's actuators, contact points, and effectors, which depend on the robot's structure and mechanical properties. This approach allows for defining multiple fitness functions that describe both design performance and control based on the condensed FEM matrices.  

We assume that a design can be characterized by the mechanical matrices of its rest state, namely $W_0$ and $\delta^{free}_0$. This hypothesis will be tested in the applications discussed in the following sections. However, $W_0$ and $\delta^{free}_0$ alone do not enable to directly generate designs. Therefore, we introduce explicit design parameters $p$, which are more intuitive for the user. The goal is to create a differentiable link as follows:

\begin{equation}
\widetilde W_0, \widetilde \delta^{\text{free}}_0 = G(p)
\label{parameter_function}
\end{equation}

where $G$ represents a MLP network. The data collection process is similar to that used for the network $F$, and the training loss follows the same approach as in Equation~\ref{eq:Loss}. Then applications such as calibration and design optimization can be explored. When design parameters are relevant, the condensed FEM model is extended as follows:

\begin{equation}
\widetilde W, \widetilde \delta^{\text{free}} = F(\delta_{a}, \delta_{c}, G(p))
\label{parameter_and_actuation_function}
\end{equation}

\begin{figure}[!ht]
\centering
\resizebox{0.5\textwidth}{!}{
\includegraphics{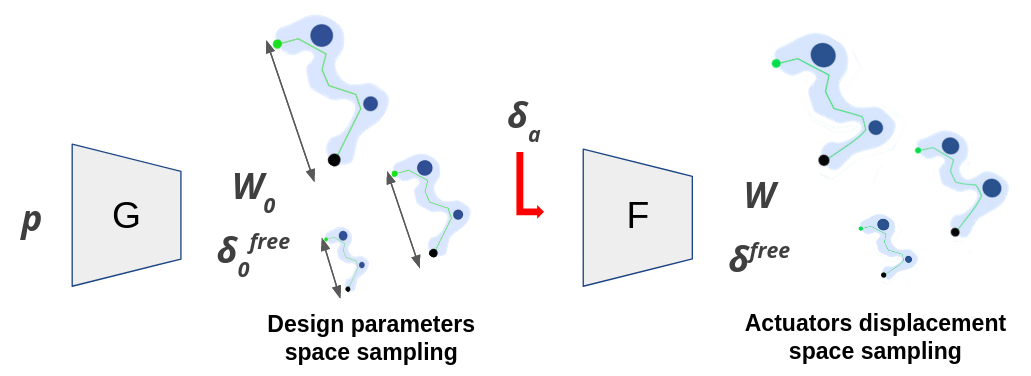}
}
\caption{Illustration of the pipeline for learning mechanical states of Soft Robot. This figure illustrates the joint use of two networks. The first network $G$ predicts the mechanical state of the robot in its rest position $(W_0, \delta^{free}_0)$ from design parameters $p$. The second network $F$ predicts the mechanical state of the robot $(W, \delta^{free})$ from both $G$ outputs and a given actuation displacement $\delta_a$.}

\label{fig:LearnedModeling}
\end{figure}

This enables building a differentiable link between design parameters and mechanical quantities, as illustrated in Figure~\ref{fig:LearnedModeling}. Differentiable fitness functions based on $W$ and $\delta^{free}$ can then be developed to evaluate design performance. 

Using the condensed FEM model for design applications offers several advantages.
In traditional design optimization processes, once a design is optimized, the simulation data are usualy discarded. While training the condensed FEM model on a data library of robot design variation still requires many simulations, once the model is trained, it can be used in both design optimization loop and control tasks without needing further simulations. Another benefit of learning both actuation states and design parameters together is that it may reduce the number of samples needed for training, compared to learning a separate condensed FEM model for each design. Additionally, since fitness functions are directly formulated from the learned matrices, a single condensed model can be used to optimize a robot for multiple tasks. 

After introducing the Soft Finger parametric design, we analyze the performance and flexibility of the condensed FEM model in relation to mesh refinement and geometric and mechanical design variables. We then explore Soft Finger design calibration and optimization scenarios, using 3 different fitness functions derived from learned mechanical matrices.

\subsection{The Soft Finger Parametric Design}

\begin{figure}[!ht]
\centering
\resizebox{0.35\textwidth}{!}{
\includegraphics{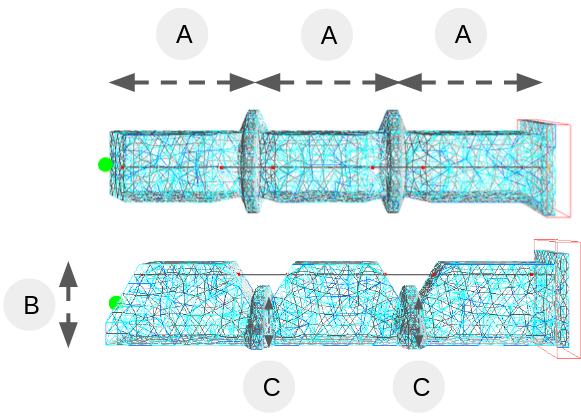}
}
\caption{Soft Finger parametric design. Design parameters are A) Length, B) Height and C) Joint Height.}
\label{fig:intro_finger_design}
\end{figure}

In our numerical experiments, a parametric design of Soft Finger is considered, based on the one proposed in~\cite{navarro2023toolbox}. Unlike the previous design in this work, this version has only one cable actuator running from the base to the tip. The parametric Soft Finger and its geometrical design variables are illustrated in Figure~\ref{fig:intro_finger_design}.

\subsection{Analysis of Mesh Refinement Influence}

\begin{figure}[!ht]
\centering
\resizebox{0.35\textwidth}{!}{
\includegraphics{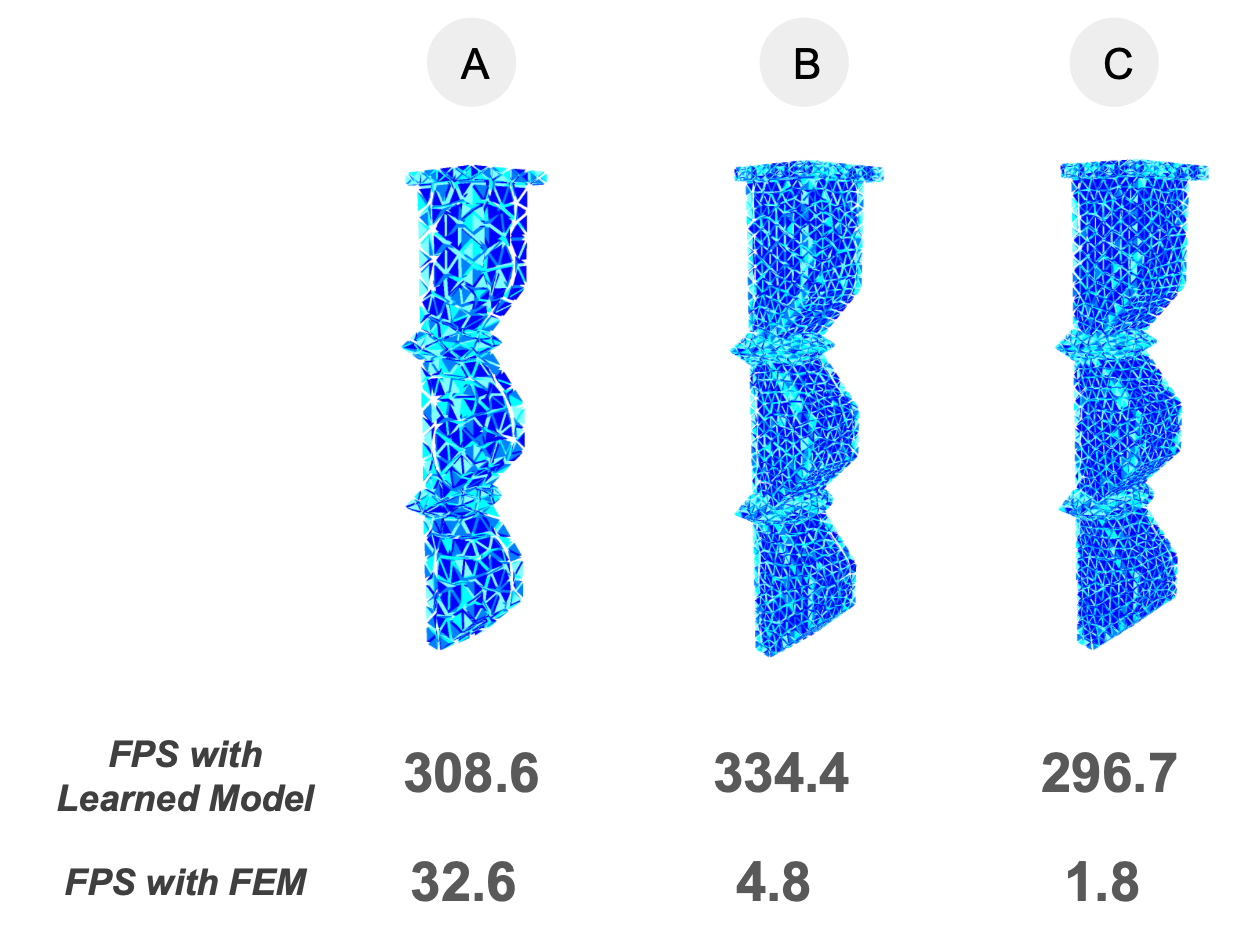}
}
\caption{Soft Fingers with different mesh resolutions: A) 1557 tetrahedrons, B) 8895 tetrahedrons, C) 17852 tetrahedrons. For each of them, the average FPS of the simulation when using the learned condensed or doing online FEM simulation is displayed.}
\label{fig:three_fingers}
\end{figure}

Mesh refinement generally leads to more accurate simulations, but increases computation time significantly. A balance between speed and accuracy is essential for real-time FEM-based control of soft robots. Since the condensed FEM model is constructed by projecting mechanical matrices into constraint space, its size remains unaffected by mesh size. When using a control method based on the learned condensed model, the extra computational costs from mesh refinement are shifted to the offline learning phase. This is illustrated by comparing three condensed FEM models of the Soft Finger with different mesh sizes (see Figure~\ref{fig:three_fingers}).

To build the training set, 6500 Soft Finger configurations are sampled, and each model is trained over 50000 episodes. The best test losses are of the same order of magnitude: $3.79 \times 10^{-6}$, $4.26 \times 10^{-6}$ and $3.60 \times 10^{-6}$ for meshes resolution A, B, and C respectively. Additionally, the FPS during the simulations remain similar across all three discretizations, as shown in Figure~\ref{fig:three_fingers}.

There is no variation in FPS when using the learned condensed model. Regardless of the mesh size, the time required for the neural network to query the mechanical state of the robot is nearly identical, resulting in similar prediction performance. The complexity of the learned condensed model does not depend on the mesh size but on the number of actuators, effectors, and contacts in the system. As mesh resolution increases, the FEM simulation is more accurate. Therefore, training on a more accurate data set leads to more reliable results without increasing the computational cost of using the trained neural network. However, this analysis does not consider contact forces, where variations in mesh size could influence contact areas and force distributions, an aspect that requires further investigation in future work.

\subsection{Parametric Design Captured by a Single Condensed FEM Model}
\label{subsec:introFingerDesignCases}

To demonstrate the expressivity of the condensed FEM model for both geometric and mechanical design variables, two condensed FEM models are trained using two different datasets with the following features. The first dataset includes 10000 Soft Finger states with varying Young's Modulus (range $[1000 kPa, 10000 kPa]$), Poisson Ratio (range $[0.400, 0.499]$), and cables displacement (range $[0.0 mm, 20.0 mm]$) sampled using a Scrambled Halton sequence. The second dataset includes 50000 Soft Finger states with varying geometrical dimensions around a baseline design of the Soft Finger as shown in Figure~\ref{fig:intro_finger_design}. The sampling spaces are 38.0mm to 42.0mm for Length, 20.0mm to 22.0mm for Height, 5.0mm to 8.0mm for Joint Height, and 0mm to 20mm for cable displacement. The open-source software Gmsh~\cite{geuzaine2009gmsh}, which enables parametric CAD through scripting, is used to generate the meshes required for simulation for each set of geometrical parameters.

This model integrates both control and design, rather than providing a separate control model for each design. Since information is shared between different designs, the dataset size is reduced compared to treating each design separately.

The training results for both datasets are shown in Table~\ref{tab:trainingDesignFingers}. A neural network with 5 fully connected layers, each with 800 nodes, is used for learning the condensed FEM model with geometrical variations of the Soft Finger. The dataset contains robot states that differ significantly from those used in previous work~\cite{Menager2023}, as the geometrical dimension parameters have a greater impact on the learned condensed FEM matrices. This results in under-sampling the training set. As an answer, a dropout strategy is applied to regularize the network. The final learning loss for the learned condensed FEM model with the geometrical dimensions' dataset, as shown in Table~\ref{tab:trainingDesignFingers}, is significantly higher than in previous cases. However, generalization remains sufficient for position-based control applications. Design variations are more difficult to learn than actuation variations. Indeed, the mechanical matrix values that link actuation to effectors, particularly in the direction influenced by changes in Length and Height, are harder to learn due to strong non-linearities. Despite this, the errors correspond to a small difference between predicted and simulated effector positions, with the worst case showing a discrepancy of around 3mm across the entire validation dataset. 

\begin{table}[!h]
    \centering
    \caption{Test loss values for each parametric Soft Finger dataset for both $G$ and $F$ neural networks.  \label{tab:trainingDesignFingers} }
    \begin{tabularx}{\linewidth}{|c||X|X|X|X|}
    \hline
     Dataset &  \multicolumn{2}{|c|}{G} & \multicolumn{2}{|c|}{F} \\ 
     \hline
      & Training Loss & Best Epoch & Final Loss & Best Epoch\\ 
      \hline
     Mechanical Parameters & 3.82e-7 & 18420 & 5.04e-5 & 12530 \\  
     \hline
     Geometrical Dimensions  & 0.19 & 4400 & 0.13 & 270 \\  
     \hline
    \end{tabularx}
\end{table}

Several designs are evaluated along the same trajectory using the condensed FEM model-based open loop inverse controller. The trajectory, defined by 25 intermediate goals, involves closing the Soft Finger on itself. Comparative trajectory errors between the positions reached by the ground truth simulation and those obtained with the condensed FEM model are shown in Table~\ref{tab:errorDesignFingers}. Regardless of the sampled design, the trained condensed FEM models can be used to control different designs from the validation datasets. In the studied examples, the maximum positioning error is around 1.2 mm, which is less than 1\% of the total length of the Soft Finger.

\begin{table}[!h]
    \centering
    \caption{Relative Euclidean positioning error for different values of geometrical and mechanical parameters.} 
    \label{tab:errorDesignFingers} 
    \begin{tabular}{|c|c|c|}
    \hline
     \textbf{Young Modulus (kPa)} & \textbf{Poisson Ratio} & \textbf{Error} \\ 
     \hline
      5000 & 0.47  & 0.91\%\\ \hline
      1000 & 0.45  & 0.29\%\\ \hline
      7000 & 0.40  & 0.90\%\\ \hline
      3000 & 0.47  & 1.03\%\\ 
     \hline
    \end{tabular}
    
    \bigskip
    \begin{tabular}{|c|c|c|c|}
    \hline
     \textbf{Length (mm)} & \textbf{Height (mm)} & \textbf{Joint Height (mm)} & \textbf{Error} \\ 
     \hline
      40.0  & 20.0  & 6.0  & 1.03\%\\ \hline
      38.5  & 21.5  & 6.0  & 3.11\%\\ \hline
      40.0  & 20.5  & 7.5  & 7.02\%\\ \hline
      41.0  & 21.0  & 5.5  & 1.62\%\\ \hline
      39.5  & 20.0  & 7.0  & 5.10\%\\
     \hline
    \end{tabular}
\end{table}

\subsection{Parameters Optimization Using a Condensed FEM Model}

\subsubsection{Application to Calibration of the Condensed FEM Model of a Parametric Soft Finger}
 
Mechanical parameters influence the robot's behavior in simulation. Since the condensed FEM model is learned from simulations, one limitation is that it is learned for a fixed set of numerical parameters. Even if the simulation is initially calibrated using a physical prototype, discrepancies may occur during the manufacturing of other similar prototypes or due to silicone degradation over time. In such cases, it would be necessary to re-calibrate the simulation and develop a new condensed FEM model for effective control. For instance, in the Soft Finger, air bubbles or incorrect dosage of silicone during the molding process can result in different mechanical parameters. To address this issue, the extended condensed FEM model is learned from the mechanical parameters' dataset. The goal is to learn the condensed FEM model with zero-shot transfer to real-world scenarios.

Once the condensed FEM model is learned, it is calibrated using data collected from the physical robot. First, the positions of the effector(s) are measured for several actuation displacement states $\delta_a$. Increasing the number of actuation states improves convergence. Next, a fitness function $O_{\text{cal}}(p)$, where $p$ represents the design parameters, is created to compute the difference between the relative position of the effectors $\delta^{*}_e$ on the physical prototype and those estimated by the condensed FEM model $\widetilde \delta_e$:
\begin{equation}
O_{\text{cal}}(p) = \sum_{\delta_a}| \widetilde \delta_e(\delta_a, p) - \delta^{*}_e(\delta_a)|
\label{eq:calibrationObjective}
\end{equation}

To build a differentiable link with the design parameters $p$, $\delta_e$ is expressed using the predicted mechanical matrices. Using Equation~\ref{eq:schur} without contacts, it becomes:
\begin{equation}
\widetilde \delta_e(\delta_a, p) = \widetilde W_{ea}(\delta_a, p) \widetilde W_{aa}^{-1}(\delta_a, p) (\delta_a - \widetilde \delta^{free}_a) + \widetilde \delta^{free}_e
\end{equation}
The design parameters $p$ are obtained by minimizing $O_{cal}$ through gradient descent.

\begin{figure}[!ht]
\centering
\resizebox{0.5\textwidth}{!}{
\includegraphics{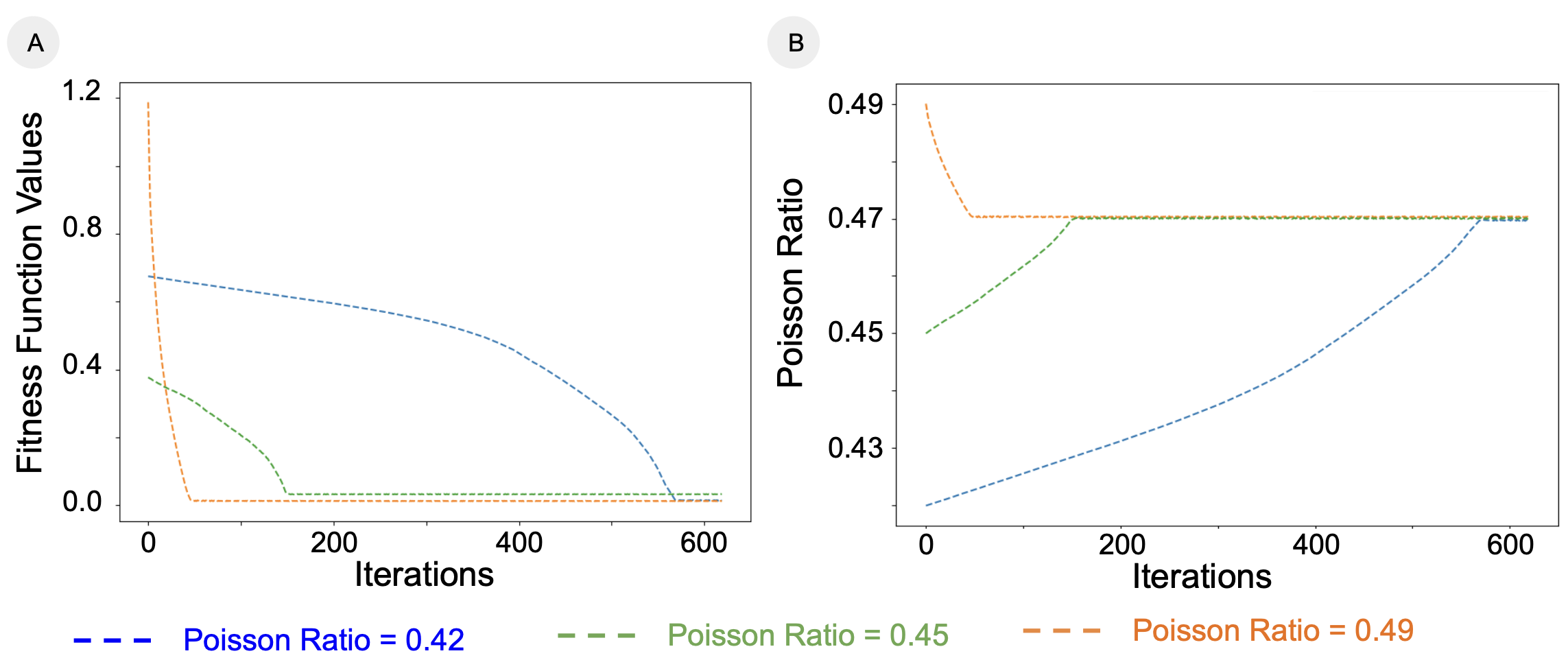}
}
\caption{Optimization history for the calibration of the Soft Finger robot. Results have been obtained with an initial learning rate of 0.01 used conjointly with an adaptive scheme as described in section~\ref{subsec:learningmodel}. Both fitness function (A) and mechanical parameters histories (B) across optimization iterations are displayed for different initial guesses of the mechanical parameters. Initial mechanical parameters are a fixed Young Modulus of 3000 kPa as well as (blue) Poisson Ratio = 0.42, (green) Poisson Ratio = 0.45, and (orange)) Poisson Ratio = 0.49. }
\label{fig:calibration_finger_different_inits}
\end{figure}

The Poisson's Ratio is optimized because the Soft Finger is controlled through displacement. In this control mode, the Poisson's Ratio is crucial for accurately modeling how the material deforms laterally in response to applied longitudinal strain. On the other hand, if the Soft Finger were controlled by applying force, the Young's Modulus would be more important, as it directly affects the stiffness and the material's resistance to deformation under load. Therefore, optimizing the Poisson's Ratio is more relevant for displacement control, while the Young's Modulus would be more critical for force control.
Since the simulation-to-reality transfer of the Soft Finger has already been validated in previous work~\cite{navarro2023toolbox}, simulation results are used as ground truth. The states $\delta^{*}_e$ of the robot are sampled in simulation for three cable displacement values $\delta_a$.
The Poisson's Ratio ground truth value is set to 0.47. The fitness function from Equation~\ref{eq:calibrationObjective} is then minimized to retrieve the corresponding parameters. The resulting convergence histories for three different initializations are shown in Figure~\ref{fig:calibration_finger_different_inits}.

In each case, the algorithm successfully finds an optimal Poisson Ratio to reduce the distance error on the effector location, with the optimization converging in just a few milliseconds. Switching to a force-based control on the cables, instead of cable displacements, would require incorporating the Young’s Modulus into the optimization process. This experiment demonstrates that it is possible to learn a single condensed FEM model that can be directly calibrated to a soft robot. By leveraging the differentiability of the learned condensed FEM model, the calibration process only takes a few seconds on a pre-learned model, unlike other approaches that rely on non-gradient solvers requiring many simulations.

\subsubsection{Application to Design Optimization of a Parametric Soft Finger}

\begin{figure}[!ht]
\centering
\resizebox{0.35\textwidth}{!}{
\includegraphics{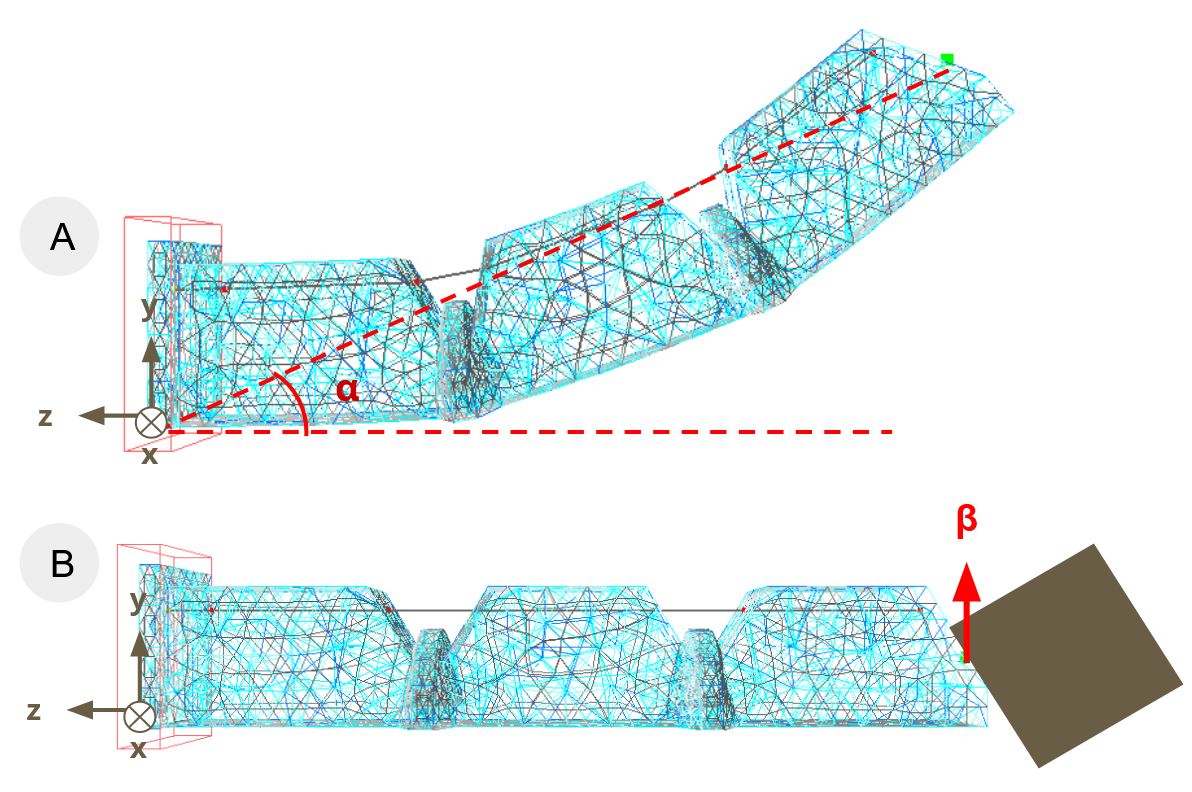}
}
\caption{Illustration of design objectives for the parametric Soft Finger. A) Bending angle $\alpha(\delta_a, p)$ reached for a fixed cable displacement $\delta_a$. B) Contact force $\beta(\delta_a, p)$ generated for a fixed cable displacement $\delta_a$.}
\label{fig:design_obj_finger}
\end{figure}

For this experiment, the parametric Soft Finger design is optimized successively for its dexterity and precision grasping. The design optimization objectives are derived from the predicted mechanical matrices. Both objectives are illustrated in Figure~\ref{fig:design_obj_finger}. 

The first fitness function $O_{dext}$ characterizes the kinematics of the parametric Soft Finger, i.e. the reached bending angle for a fixed cable displacement $\delta_a = 10$ mm. Minimizing this function is equivalent to maximizing the workspace of the robot. $O_{dext}$ is expressed as follows:
\begin{equation}
\begin{aligned}
O_{\text{dext}}(p) &= |\alpha^{\text{max}} - \alpha(\delta_a, p)| \\
\alpha(\delta_a, p) &= Arccos(\frac{\widetilde \delta^z_e(\delta_a, p)}{|\delta_e(\boldsymbol{0}, p)|})
\end{aligned}
\label{eq:Angular objective}
\end{equation}
where $\alpha$ represents the function that gives the bending angle in the z-plane and $\alpha^{\text{max}} = 1.57$ rad is the maximum achievable angle. 

The second fitness function $O_{str}$ measures the contact force generated at the tip of the Soft Finger for a fixed cable displacement $\delta_a = 10$ mm. Minimizing this function maximizes the performance of the Soft Finger for precision grasping. Again, to establish a differentiable link with the design parameters $p$, the contact force $\lambda_c$ is derived from the predicted mechanical matrices. Assuming a fixed object (i.e. $\delta_c = 0$) and considering the tip effector as a contact point, the system expressed through both $\delta_a$ and $\delta_c$ from Equation~\ref{eq:schur} gives: 
\begin{equation}
\lambda_c(p,\delta_a) = B(p,\delta_a)^{-1} [\tilde W_{ca} \tilde W_{aa}^{-1} (\delta_a - \tilde \delta_a^{\text{free}}) + \tilde \delta_c^{\text{free}}]
\label{eq:lambdac_parametric_finger}
\end{equation}
with $B(p,\delta_a) = \tilde W_{ca} \tilde W_{aa}^{-1} \tilde W_{ac} - \tilde W_{cc}$
Finally, the fitness function $O_{str}$ is expressed as follows:
\begin{equation}
\begin{aligned}
O_{\text{str}}(p) &= |\beta^{max} - \beta(\delta_a, p)| \\
\beta(\delta_a, p) &= |\lambda_c(p,\delta_a)^y|
\end{aligned}
\label{eq:Angular objective}
\end{equation}
where $\beta$ is the function giving the force generated at the tip of robot on the y-axis and $\beta^{max} = 10000$ N is an unreachable force. 

The two fitness functions are formulated using the mechanical matrices learned with the condensed FEM model. These matrices link forces to displacements in key areas of the design while incorporating its mechanical structure. This enables the creation of fitness functions from this compact representation. As a result, design optimization can leverage this pre-learned model to efficiently explore the design space. In the following, results are first generated and analyzed separately for each fitness function. Then, we demonstrate how a multi-objective optimization scheme can be implemented using the condensed FEM modeling.\\

\textbf{Single objective optimization:}  Since the considered fitness functions are highly non-convex, a gradient descent optimizer may fall in a local optimum. To avoid this, a grid search is first performed to identify a good starting point for optimization. As the design space is three-dimensional, this computation takes only a few seconds with the condensed FEM model. A total of 600 designs are evaluated for both fitness functions. In higher-dimensional cases, stochastic gradient descent could be used instead of standard gradient descent to mitigate dependency on the initialization.

\begin{figure}[!ht]
\centering
\resizebox{0.46\textwidth}{!}{
\includegraphics{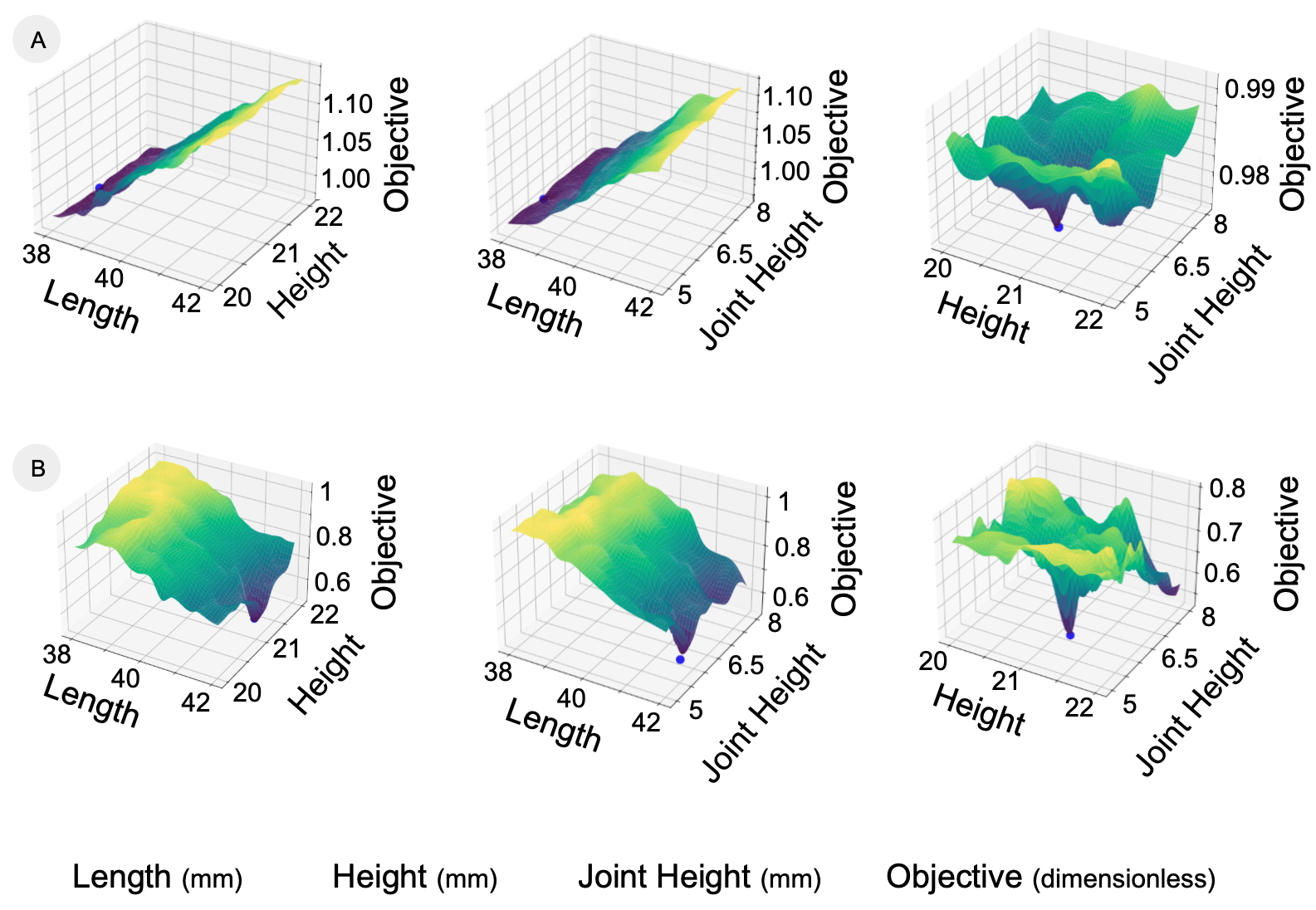}
}
\caption{Fitness landscapes around the initialized design selected with grid search strategy, for A) dexterity fitness function (Length = 38.0mm, Height = 20.57mm, Joint Height = 6.28mm) and B) strength fitness function (Length = 42mm, Height = 22mm, Joint Height = 8mm). The best design after optimization is shown as a blue dot.}
\label{fig:fitness_landscapes_single_obj}
\end{figure}

For each fitness function, the objective function landscape is computed using a regular grid sampling strategy around the chosen initial design. The resulting landscapes for both fitness functions are shown in Figure~\ref{fig:fitness_landscapes_single_obj}, revealing that Length has a greater influence on both objectives compared to the other two design variables.

Gradient descent is performed starting from the initial design. For dexterity optimization, the best design (Length=38.0 mm, Height=20.6mm, Joint Height=6.3mm) favored a shorter length. In contrast, for strength optimization, a longer length  (Length=42.0 mm, Height=22.0mm, Joint Height=7.9mm) helps generate more force on the object due to the lever arm effect. The best design for maximizing contact forces at the tip also featured a Soft Finger with more material, providing better rigidity around the contact point. \\

\begin{figure}[!ht]
\centering
\resizebox{0.37\textwidth}{!}{
\includegraphics{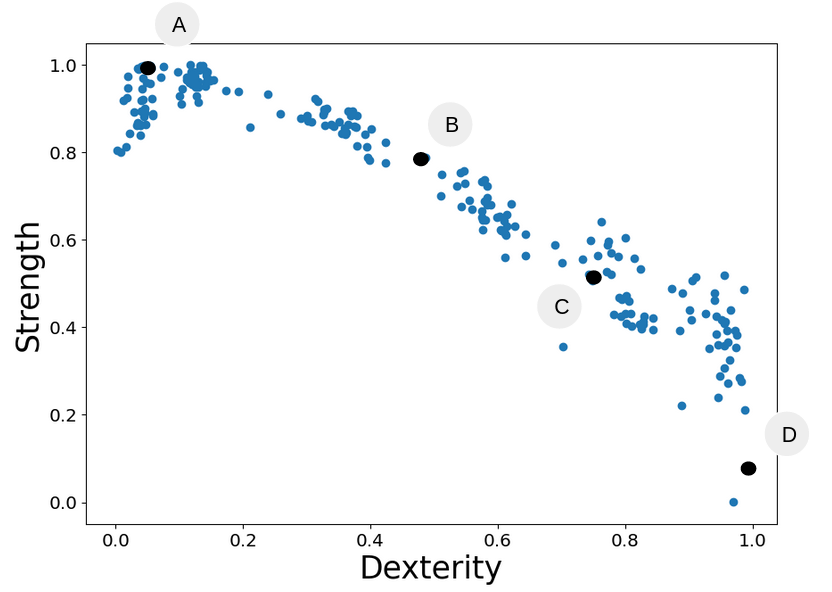}
}
\caption{Pareto Front for the design optimization of the Soft Finger regarding Dexterity (maximization of the bending angle for a given actuation state) and Strength (maximization of the contact force for a given actuation state) metrics and generated using a Grid Search strategy. These two metrics are normalized and therefore dimensionless. Design parameters for several designs sampled: A) Length = 38.0mm, Height = 21.6mm, Joint Height = 5.0mm, B) Length = 40.4mm, Height = 20.0mm, Joint Height = 5.6mm, C) Length = 40.2mm, Height = 22.0mm, Joint Height = 7.4mm, D) Length = 42.0mm, Height = 22.0mm, Joint Height = 8.0mm}
\label{fig:multi_obj_pareto}
\end{figure}

\textbf{Multi-objective optimization:} In this experiment, we highlight how to determine the optimal design of a Soft Finger based on several metrics computed using the condensed FEM model. Figure \ref{fig:multi_obj_pareto} shows the fitness functions values for 600 Soft Finger designs sampled with a grid search strategy. To allow for comparison between the two losses, both are normalized using $O^{\text{norm}}_i(p) = \frac{O_i(p) - \text{min}_p*(O_i(p*))}{\text{max}_p*(O_i(p*)) - \text{min}_p*(O_i(p*))}$ where $p*$ represents the design parameters encountered during initial grid search. 

A Pareto front is visible, indicating that the two considered metrics are antagonistic. Therefore, a compromise has to be found. This can be achieved by using the gradients calculated through the condensed model and formulating a single fitness function that aggregates the two metrics:
\begin{equation}
O_{\text{mult}}(p) = \gamma_1 O^{\text{norm}}_{\text{dext}}(p) + \gamma_2 O^{\text{norm}}_{\text{str}}(p) 
\label{eq:lambdac_parametric_finger}
\end{equation}
where both $\gamma_1$ and $\gamma_1$ are user-defined weighting factors between 0 and 1, allowing users to prioritize each fitness function. This approach enables the generation of the optimal design of a soft robot based on user-specific requirements. Setting these weighting factors is typically challenging and often done by trial and error, with each trial requiring numerous simulations to evaluate the fitness functions at each optimization step. Using a pre-learned condensed FEM model eliminates the computational time associated with these simulations.

\section{Conclusion And Perspectives}
 
This work extends the condensed FEM model initially introduced in~\cite{Menager2023}. Below, we summarize the main contributions, limitations, and potential future directions for exploration.

First, contact modeling has been integrated into the learned condensed FEM model framework. Contacts are treated as constraints, similar to actuators and effectors. We demonstrate how to formulate inverse control problems for both single and coupled soft robots. In the current framework, we assumed that the number and location of the contact points are fixed. This limitation could be addressed by considering multi-point contact, using models capable of predicting variable-sized matrices, such as Transformers~\cite{Transformersc2017} or GNN~\cite{GNN}, or by expanding the robot's state to include multiple potential contact points.

Secondly, the embedding of the condensed FEM model for both open-loop and closed-loop control is demonstrated on a pneumatic soft robot. Once learned, we show that the model can be deployed on an embedded controller. The embedded control could be applied to other robots. The more complex the robot, the greater the potential speed-up. However, significant changes in actuation or morphology would require re-learning the model. So far, we have only considered a quasi-static assumption. This limitation could be addressed by considering learning the dynamics. Neural architectures designed for sequential data, such as LSTMs\cite{LSTM1997} or Transformers~\cite{Transformersc2017}, are promising candidates for this task. Extending the condensed FEM model to include dynamics could enable the use of more robust control methods, such as Model Predictive Control \cite{MPC1989}. 

In this article, we demonstrate that the condensed FEM model formalism can be used to jointly learn both the design and control of a soft robot. Applications in direct calibration of the condensed FEM model and design optimization are showcased using the example of a parametric Soft Finger. So far, our model has been tested with a limited number of design parameters. Expanding this approach to more complex parameterization, with a greater number of parameters or increased expressivity, is an interesting prospect that would require rethinking network scaling and sampling strategies. Moreover, this type of model could provide an ideal foundation for developing co-optimization techniques. Currently, the proposed control is low-level, based on inverse robot modeling and quadratic programming optimization. However, it is conceivable to use the learned condensed FEM model as a surrogate model in a RL approach, where the control strategy incorporates design information to solve high-level control tasks and optimize the robot's design simultaneously. Another limitation of the method is its dependence on fixed material and mechanical parameters, which may change over time or across different robots. However, this can be mitigated using domain randomization, as a single network can be used with different geometric or mechanical properties.

Finally, to address the limitation of the offline sampling process, one potential extension is to use online learning techniques. Indeed, one of the drawbacks of the method is its reliance on a large amount of offline data. However, many robot states may be redundant or irrelevant to the specific task at hand. Online learning would allow for case-dependent exploration of the actuation space, which is particularly useful when multiple contact points are involved, as this leads to a significantly larger sampling space. Moreover, in the case of very dense sampling, the learned function primarily serves an interpolation role rather than extrapolation. The challenge of learning out-of-distribution behavior is an important question that will need to be addressed in future work. Although we consider entirely soft robots, the current condensed formalism also supports both rigid and soft components. Additionally, Physics Informed Neural Network~\cite{PINNS2021} could be implemented to enhance learning generalization. This would require reformulating the problem to incorporate first-order derivatives for the learned variables.

\vfill

\end{document}